%% file: main.tex
\documentclass{ecai} 

\input{header}

\newcommand{\BibTeX}{B\kern-.05em{\sc i\kern-.025em b}\kern-.08em\TeX}

\begin{document}


\begin{frontmatter}


\paperid{2412} 


\title{Reset It and Forget It: Relearning Last-Layer Weights Improves Continual and Transfer Learning}


\author[A]{\fnms{Lapo}~\snm{Frati}\footnote{Equal contribution.}\footnote{Corresponding authors. Email: \{lfrati,ntraft\}@uvm.edu.}}
\author[A]{\fnms{Neil}~\snm{Traft}\footnotemark[1]\footnotemark[2]}
\author[B,C,D]{\fnms{Jeff}~\snm{Clune}} 
\author[A]{\fnms{Nick}~\snm{Cheney}} 

\address[A]{University of Vermont}
\address[B]{University of British Columbia}
\address[C]{Vector Institute}
\address[D]{Canada CIFAR AI Chair}


\begin{abstract}
This work identifies a simple pre-training mechanism that leads to representations exhibiting better continual and transfer learning. This mechanism---the repeated resetting of weights in the last layer, which we nickname ``{\zapping}''---was originally designed for a meta-continual-learning procedure, yet we show it is surprisingly applicable in many settings beyond both meta-learning and continual learning. In our experiments, we wish to transfer a pre-trained image classifier to a new set of classes, in few shots. We show that our {\zapping} procedure results in improved transfer accuracy and/or more rapid adaptation in both standard fine-tuning and continual learning settings, while being simple to implement and computationally efficient. In many cases, we achieve performance on par with state of the art meta-learning without needing the expensive higher-order gradients by using a combination of {\zapping} and sequential learning. An intuitive explanation for the effectiveness of this {\zapping} procedure is that representations trained with repeated {\zapping} learn features that are capable of rapidly adapting to newly initialized classifiers.  Such an approach may be considered a computationally cheaper type of, or alternative to, meta-learning rapidly adaptable features with higher-order gradients.  This adds to recent work on the usefulness of resetting neural network parameters during training, and invites further investigation of this mechanism.
\end{abstract}

\end{frontmatter}


\section{Introduction}

Biological creatures display astounding robustness, adaptability, and sample efficiency; while artificial systems suffer ``catastrophic forgetting'' \citep{mccloskey1989catastrophic, french1999catastrophic} or struggle to generalize far beyond the distribution of their training examples.  It has been observed in biological systems that repeated exposure to stressors can result in the evolution of more robust phenotypes \citep{rohlf2009emergent, felix2008robustness}.  However, it is not clear what type of stressor during the training of a neural network would most effectively and efficiently convey robustness and adaptability to that system at test time.  

\begin{figure}[!ht]
    \centering
    \includegraphics[width=0.97\linewidth]{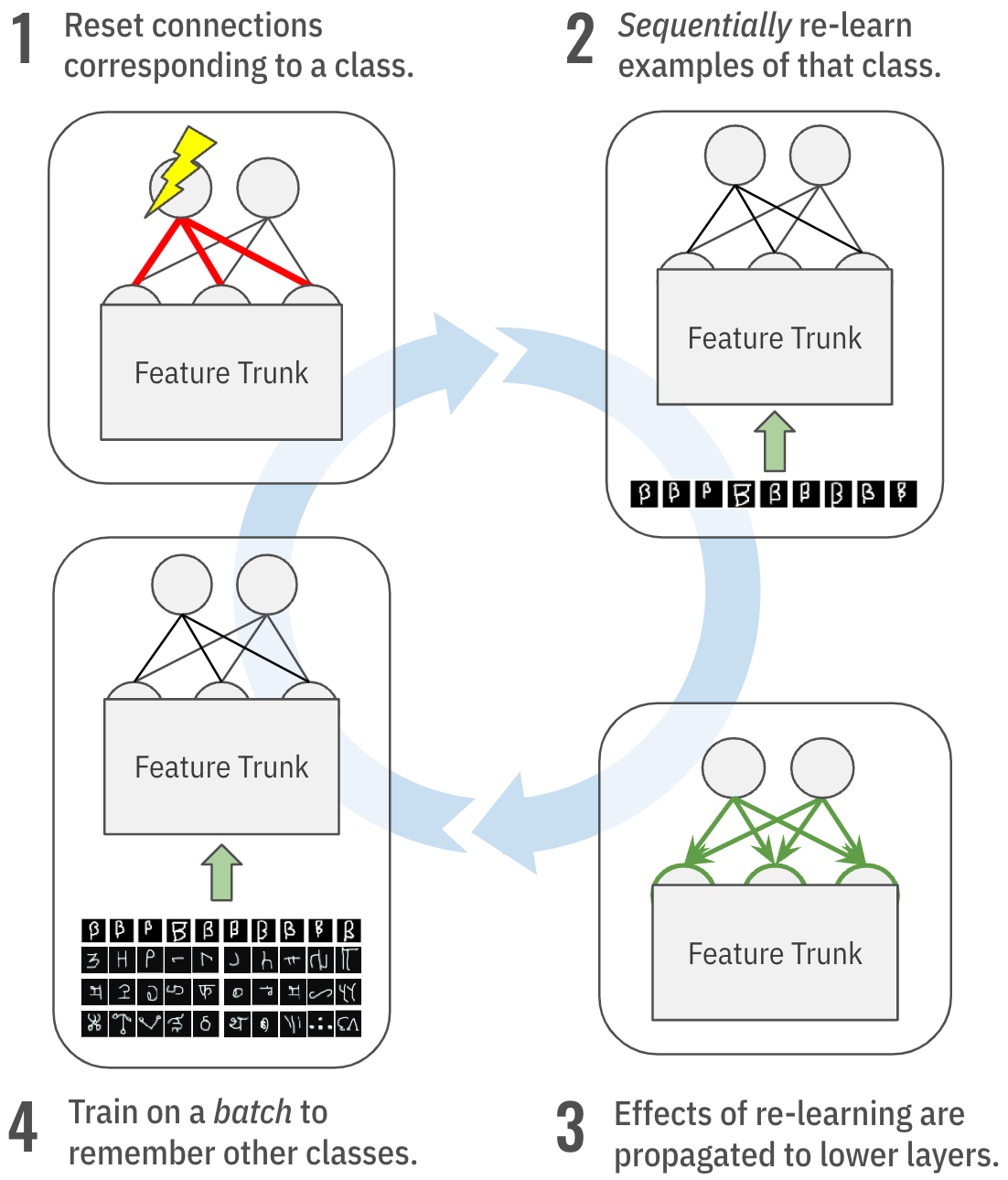}
    \vspace{1em}
    \caption{\textit{Alternating Sequential and Batch} learning (ASB) alternates between phases of (Step 2) individual examples from a single class, and (Step 4) multi-class batches of examples. Before each sequential phase the existing class is forgotten by {\color{egg}\Large\Lightning} {\zapping}.}
    \vspace{2em}
    \label{fig:zap_toon}
\end{figure}

It is common practice to have the training of a machine learning system mimic the desired use-cases at test time as closely as possible. In the case of an image classifier, this would include drawing independent training samples from a distribution identical to the test set (i.i.d.\ training). When the test scenario is \textit{itself} a learning process---such as few-shot transfer learning from a limited number of novel examples---training can include repeated episodes of rapid adaptation to small subsets of the whole dataset, thereby mimicking the test scenario.  Meta-learning algorithms (also known as learning-to-learn) for such contexts are able to identify patterns that generalize across individual learning episodes \citep{schmidhuber1987evolutionary, bengio1990learning, hochreiter2001learning, ravi2017optimization, finn2017model}. 

Going even further, we are interested in challenging scenarios of \textit{online/continual learning with few examples per class}. We ask, if this is the test scenario, then what kind of training would best mimic this? What kind of stressor could be applied \textit{in pre-training} to confer robustness \textit{upon deployment} into the difficult setting of few-shot continual learning?

Recent work by \citet{javed2019meta} and \citet{beaulieu2020learning} has tackled this question with the application of meta-learning.
When the learning process itself is differentiable, one way to perform meta-optimization is by differentiating through several model updates using second-order gradients, as is done in Model-Agnostic Meta-Learning (MAML) \citep{finn2017model}.
When applied to episodes of continual learning, this approach induces an implicit regularization (it penalizes weight changes that are disruptive to previously learned concepts, without needing a separate heuristic to quantify the disruption).

The authors of Online-Aware Meta-Learning ({\oml}) \citep{javed2019meta} divide their architecture into two parts, where later layers are updated in a fast inner loop but earlier layers are only updated in a slower outer meta-update.
Subsequently, A Neuromodulated Meta-Learning Algorithm ({\anml}) \citep{beaulieu2020learning} restructured the {\oml} setup by moving the earlier meta-only layers into a parallel neuromodulatory path which meta-learned an attention-like context-dependent multiplicative gating, achieving state of the art performance in sequential transfer.

However, in this work we show that neither the asymmetric training of different parts of the network (like OML), nor a neuromodulatory path (like ANML), nor even expensive second-order updates (like both), are necessary to achieve equivalent performance in this setting. 
Instead, we reveal that the key contribution of the OML-objective appears to be a previously overlooked 
mechanism: \textit{a weight resetting and relearning procedure}, which we call ``{\zapping}''.
This procedure consists of frequent random resampling of the weights leading into one or more output nodes of a deep neural network, which we refer to as \textquote{\zapping} those neurons.%
\footnote{In OML, weight resampling was employed primarily as a method for maintaining large meta-gradients throughout meta-training and was deemed non-essential (\citet[arXiv Version 1, Appendix A.1: Random Reinitialization]{javed2019meta}).} 

After reinitializing the last layer weights, the gradients for the weights in upstream layers quantify how the representation should have been different to reduce the loss \textit{given a new set of random weights}. This is exactly the situation that the representation will be presented with during transfer. 
In the case where all weights of the last layer are reset, {\zapping} closely matches the common technique of transfer learning by resetting the classifier layer(s) on top of pre-trained feature extraction layers (as in \citet{yosinski2014transferable}). 
Over multiple repetitions, this leads to representations which are more suitable for transfer.

We are able to show the efficacy of this forgetting-and-relearning by introducing a variation which rehearses a continual learning process \textit{without} meta-learning. We call this \textit{Alternating Sequential and Batch} learning (ASB, Figure \ref{fig:zap_toon}). This process runs through episodes of continual learning just like {\oml}/{\anml}, but does not perform meta-optimization. However, it does zap a neuron at the beginning of each continual learning episode, giving the model a chance to relearn the forgotten class. We hypothesize that this better prepares the model for a similar learning process at transfer time.

We show that:
\begin{itemize}
  \item The \textit{\zapping} forget-and-relearn mechanism accounts for the majority of the meta-learning improvements observed in prior work (Section~\ref{sec:olft}).
  
  \item Dedicating second-order optimization paths for certain layers doesn't explain the performance of meta-learning for continual transfer (Section~\ref{sec:olft}).

  \item \textit{Alternating Sequential and Batch} (ASB), with {\zapping}, is often sufficient to match or outperform meta-learning without any expensive bi-level optimization (Sections~\ref{sec:olft} and~\ref{sec:iid-transfer}).
  
  \item Representations learned by models pre-trained using {\zapping} are better for general transfer learning, not just continual learning (Section~\ref{sec:iid-transfer}).
  
  \item {\Zapping} and ASB can be useful across different datasets and model architectures (Section~\ref{sec:bigger_is_better}).

\end{itemize}

Source code for our methods are available at:\\
\url{github.com/uvm-neurobotics-lab/reset-it-and-forget-it}

\section{Methods} \label{sec:methods}


As described in \citet{javed2019meta} and \citet{beaulieu2020learning}, we seek to train a model capable of learning a large number of tasks $\mathcal{T}_{1..n}$, in a few shots per task, with many model updates occurring as tasks are presented \emph{sequentially}. Tasks $\mathcal{T}_{i}$ come from a common domain $\mathcal{D}$. In our experiments we consider the domain to be a natural images dataset and tasks to be individual classes $\mathcal{C}_i$ in that dataset.
 
Recent works show that reinitialization of some layers during training can be used as a regularization method \citep{zhao2018retraining, li2020rifle, alabdulmohsin2021impact, zhou2022fortuitous, chen2023improving}. But there are several ways in which this reinitialization can be applied.
We focus our attention on the last fully connected layer, right before the output. While the information value within this last layer may be marginal \citep{hoffer2018fix, zhang2022all} interventions in the last layer will affect the gradient calculation of all the other layers in the network during backpropagation (Figure~\ref{fig:zap_toon}, Step 3).

For consistency with prior approaches, our primary experiments employ the model architecture from~\citet{beaulieu2020learning} (minus the neuromodulatory path). This model consists of a small convolutional network with two main parts: a stack of convolutional layers that act as a feature extractor, and a single fully connected layer that acts as a linear classifier using the extracted features (see Appendix~\ref{si:convnets}, Figure~\ref{fig:si-convnet}). We then extend our findings on the larger VGG architecture \citep{simonyan2014very} to demonstrate the scalability of our approach.

Our ``{\zapping}'' procedure consists of re-sampling all the connections corresponding to one of the output classes.
Because of the targeted disruption of the weights, the model suddenly forgets how to map the features extracted by previous convolutional layers to the correct class. To recover from this sudden change, the model is shown examples from the forgotten class, one at a time. By taking several optimization steps, the model recovers from the negative effect of {\zapping}. This procedure constitutes the \textit{inner loop} in our meta-learning setup {and is followed by an \textit{outer loop} update using examples from all classes~\citep{javed2019meta}. The outer loop update is done with a higher-order gradient w.r.t. the initial weights of that inner loop iteration \citep{finn2017model}.} But, as we will see, these inner and outer updates do not actually need to be performed in a nested loop---they can also be arranged in a flat sequence without meta-gradients, yielding similar performance with much more efficient training.

\subsection{Training Phases} \label{sec:training}


Since we want to learn each class in just a few shots, it behooves us to start with a pre-trained model rather than starting tabula rasa. Therefore our set-up involves two stages. Within each stage, we examine multiple possible configuration options, described in more detail in the next sections.

\begin{enumerate}
    \item (Sec.~\ref{sec:pre-training}) \textbf{Pre-Training:} We use one of the following algorithms to train on a subset of classes:
    (1)~standard \textit{i.i.d.}\ pre-training,
    (2)~alternating sequential and batch learning (\textit{ASB}),
    or (3)~meta-learning through sequential and batch learning (\textit{meta-ASB}). 
    Each of these may or may not include the \textbf{\zapping} procedure to forget and relearn.  
    \item (Sec.~\ref{sec:evaluation}) \textbf{Transfer:} Following pre-training, we transfer the classifier to a separate subset of classes using (1)~\textit{sequential transfer} (continual learning) or (2)~standard \textit{i.i.d.\ transfer} (fine-tuning).
\end{enumerate}

\subsubsection{Stage 1: Pre-Training}
\label{sec:pre-training}

Our pre-training algorithm is described in Algorithm~\ref{alg:pretraining}, and visualized in Appendix~\ref{si:alternating}, Figure~\ref{fig:si-alternating}.  Our algorithm is based on the \textquote{Online-aware} Meta-Learning ({\oml}, \citet{javed2019meta}) procedure, which consists of two main components:

\begin{itemize}
  \item \textbf{Adapting} (\textit{inner loop; sequential learning}): In this phase the learner is sequentially shown a set of examples from a single random class and trained using standard SGD. The examples are shown one at a time, so that the optimization performs one SGD step per image.
  \item \textbf{Remembering} (\textit{outer loop; batch learning}): After each adapting phase, the most recent class plus a random sample from all classes are used to perform a single outer-loop batch update. Those samples serve as a proxy of the true meta-loss (learning new things without forgetting anything already learned). The gradient update is taken w.r.t.\ the initial inner-loop parameters, and those updated initial weights are then used to begin the next inner-loop adaptation phase following the MAML paradigm \citep{finn2017model}.
\end{itemize}

Compared to {\oml} we use a different neural architecture (which improves classification performance; Appendix~\ref{si:convnets}, Figure~\ref{fig:si-convnet}), and do not draw a distinction between when to update the feature extraction vs.\ classification layers (updating both in the inner and outer loops).
Furthermore, while the original {\oml} procedure included both zapping and higher-order gradients, we ablate the effect of each component by allowing them to be turned on/off as follows.

In configurations with \textbf{\zapping} (denoted as {\zap} in Algorithm~\ref{alg:pretraining}), prior to each sequential adaption phase on a single class $\mathcal{C}_i$, the final layer weights corresponding to that class are re-initialized---in other words, they are re-sampled from the initial weight distribution\footnote{Weights are sampled from the Kaiming Normal \citep{he2015delving} and biases are set to zero.}. We call this procedure {\zapping} as it destroys previously learned knowledge that was stored in those connections.

\begin{algorithm}[t!]
    \caption{Pre-Training: ASB and Meta-ASB, with or without {\zapping}} 
    \label{alg:pretraining}
    \begin{algorithmic}[1]
    \REQUIRE Dataset $\mathcal{D} :$ C classes, N examples per class, (H, W, Ch) images
    \REQUIRE Network $f :$ (H, W, Ch) $\rightarrow \mathcal{C}$ with parameters $\theta : [\theta^{\color{blue}conv},\theta^{\color{orange}fc}]$ 
    \REQUIRE $\eta_{\text{in}},\;\eta_{\text{out}}$ inner and outer learning rates 
    \REQUIRE $K$ number of sequential inner-loop examples
    \REQUIRE $R$ number of outer-loop ``remember'' examples
    \REQUIRE $S$ number of outer-loop steps
    
    \FOR[\textbf{outer loop; remembering}]{iteration $=1,2,\ldots,S$}
    
    \STATE $\mathcal{C} \sim \mathcal{D}$
    \COMMENT{{\scriptsize \color{gray} Sample one class}}
    \STATE $X_{\text{inner}} \sim \mathcal{C}$
    \COMMENT{{\scriptsize \color{gray} $K$ examples from class $\mathcal{C}$}}
    \STATE $X_{\text{rand}} \sim \mathcal{D}$
    \COMMENT{{\scriptsize \color{gray} $R$ examples from the whole dataset}}
    \STATE $X_{\text{outer}} \leftarrow X_{\text{inner}} \cup X_{\text{rand}}$
    
    \IF{\zap} \label{alg:line:zap}
        \STATE Reset connections in $\theta^{\color{orange}fc}$ corresponding to class $\mathcal{C}$ \COMMENT{\textbf{\zapping}\label{zapping}}
    \ENDIF
    
    \STATE $\theta_0 \leftarrow \theta$
    \FOR[\textbf{inner loop; adapting}]{i = 0, \ldots , K-1} \label{alg:line:seq}
        \STATE $\hat y \leftarrow f(X_{\text{inner}}^{i}; \theta_{i})$
        \STATE $\theta_{i+1} \leftarrow \theta_{i} - \eta_{\text{in}} \nabla_{\theta_{i}}\mathcal{L}(\hat y, \mathcal{C})$ \COMMENT{{\scriptsize \color{gray} single example SGD}}
    \ENDFOR
    
    \IF{\meta} \label{alg:line:meta}
        \STATE $\theta \leftarrow {\color{red}\theta_0} - \eta_{\text{out}} \nabla_{{\color{red}\theta_0}}\mathcal{L}(f(X_{\text{outer}}; \theta_{K}), Y)$ \COMMENT{{\scriptsize \color{gray} meta batch SGD (expensive)}}
    \ELSE
        \STATE $\theta \leftarrow {\color{red}\theta_{K}} - \eta_{\text{out}} \nabla_{{\color{red}\theta_{\text{K}}}}\mathcal{L}(f(X_{\text{outer}}; \theta_{K}), Y)$ \COMMENT{{\scriptsize \color{gray} standard batch SGD (cheap)}}
    \ENDIF
    
    \ENDFOR
    \end{algorithmic} 
\end{algorithm}

In the \textbf{meta-learning} conditions (denoted as {\meta} in Algorithm~\ref{alg:pretraining}), the \textquote{remembering} update is performed as an \textit{outer}-loop meta-update w.r.t. the initial weights of each inner-loop (as described above), and these meta-updated weights are then used as the starting point of the next inner-loop. However, we also wish to examine the effect of {\zapping} independent of meta-learning, and introduce a new pre-training scenario in which we \textit{alternate} between the adapting phase and the remembering phase. Different from meta-learning, this new method does not backpropagate through the learning process nor rewind the model to the beginning of the inner loop. Instead, it simply takes normal (non-meta) gradient update steps for each image/batch seen. The weights at the end of each sequence-and-batch are used directly on the next sequence.  

We refer to this approach as \textit{Alternating Sequential and Batch} learning (\textbf{ASB}), and the difference between ASB and meta-ASB can be seen visually in App.~\ref{si:alternating}, Fig.~\ref{fig:si-alternating}. This approach---like Lamarckian inheritance rather than Darwinian evolution \citep{fernando2018meta}---benefits from not throwing away updates within each inner-loop adaptation sequence, but loses the higher-order updates thought to be effective for these continual learning tasks~\citep{javed2019meta, beaulieu2020learning}. This sequential approach allows us to employ the same {\zapping} procedure as above: resetting the output node of the class which we are about to see a sequence of.

We also wished to study how {\zapping} may influence the learning of generalizable features without being coupled with sequential learning. Thus, we also apply {\zapping} to \textbf{i.i.d.\ pre-training}, which uses standard mini-batch learning with stochastic gradient descent. 
In this setting, a random sample of classes are zapped at a configurable cadence throughout training. For example, we might resample the entire final layer once per epoch, allowing the network to experience an event which is similar to fine-tuning.

\subsubsection{Stage 2: Transfer} \label{sec:evaluation}

We evaluate our pre-trained models using two different types of few-shot transfer learning. 

In \textbf{sequential transfer} (Alg.~\ref{alg:eval_seq}), we follow the evaluation method used in prior works \citep{javed2019meta,beaulieu2020learning}. The model is trained on a long sequence of different unseen classes (\textit{continual learning}). Examples are shown one at a time, and a gradient update is performed for each image. Only the weights in the final layer are updated (also referred to as ``linear probing'' \citep{alain2016understanding}).

We also test \textbf{i.i.d.\ transfer} (Alg.~\ref{alg:eval_iid}), where the model is trained on batches of images randomly sampled from unseen classes (\textit{standard fine-tuning}). 

In both transfer scenarios, the new classes were not seen during the pre-training phase. There are only 15-30 images per class (\textit{few-shot learning}). Between the end of pre-training and transfer, the final linear layer of the model is replaced with a new, randomly initialized linear layer, so it can learn a new mapping of features to these new classes. 
Both sequential and i.i.d. transfer use the same set of classes and images---the only difference is how they are presented to the model.

\begin{algorithm} [ht]
\caption{Sequential Transfer Protocol\label{alg:eval_seq} (adapted from \citet{beaulieu2020learning}, Algorithm 2)} 
\begin{algorithmic}[1]
    \REQUIRE $\mathcal{C} \gets$ sequential trajectory of $N$ unseen classes
    \REQUIRE $\theta \gets $ pre-trained weights of the network
    \REQUIRE $\beta \gets $ learning rate hyperparameter
    
    \STATE $S_{train} = [\ ] $ 
    \FOR{$n=1,2, \ldots,N$}  
        \STATE $S_{traj} \sim \mathcal{C}_n$ \COMMENT{{\scriptsize \color{gray} get training examples from next class}}
        \STATE $S_{train} = S_{train} \bigcup S_{traj}$ \COMMENT{{\scriptsize \color{gray} add to seq. transfer train set}}
         \FOR{$i=1,2, \ldots,k$}      
            \STATE $\theta \gets \theta - \beta \nabla_{\theta}\mathcal{L}(\theta,S^{(k)}_{traj})$
            \COMMENT{{\scriptsize \color{gray} SGD update on a single image}}
        \ENDFOR
    
        \STATE record $\mathcal{L}(\theta,S_{train}) $
        \COMMENT{{\scriptsize \color{gray} eval current $\theta$ on classes trained so far}}
        \STATE $ S_{test} = ( \bigcup\limits_{1 \ldots n} \mathcal{C}_{i} ) - S_{train} $ \COMMENT{{\scriptsize \color{gray} held-out from seen classes}}
        \STATE record $\mathcal{L}(\theta,S_{test}) $
        \COMMENT{{\scriptsize \color{gray} eval current $\theta$ on held-out examples}}
    
        \ENDFOR

\end{algorithmic}
\end{algorithm}

\begin{algorithm} [ht]
\caption{I.I.D. Transfer Protocol\label{alg:eval_iid}} 
\begin{algorithmic}[1]
    \REQUIRE $\mathcal{D}_{tr}, \mathcal{D}_{te} \gets$ training and held-out examples from $N$ unseen classes from domain $\mathcal{D}$
    \REQUIRE $\theta \gets $ pre-trained weights of the network
    \REQUIRE $\beta \gets $ learning rate hyperparameter
    \REQUIRE $E \gets $ number of training epochs
    
    \FOR{$i=1,2, \ldots, E$}
    \FOR[{\scriptsize \color{gray} $N$ is the number of batches in} $\text{\scriptsize \color{gray} $\mathcal{D}_{tr}$}$]{$i=1,2, \ldots, N$}
        \STATE $B_{i} \sim \mathcal{D}_{tr}$ \COMMENT{{\scriptsize \color{gray} uniformly sample from $\mathcal{D}_{tr}$ w/o replacement}}
        \STATE $\theta \gets \theta - \beta \nabla_{\theta}\mathcal{L}(\theta,\mathcal{B}_i)$
        \COMMENT{{\scriptsize \color{gray} standard batch SGD update}}
        
        \STATE record $\mathcal{L}(\theta, \mathcal{D}_{tr}) $
        \COMMENT{{\scriptsize \color{gray} eval current $\theta$ on all classes}}
        \STATE record $\mathcal{L}(\theta, \mathcal{D}_{te}) $
        \COMMENT{{\scriptsize \color{gray} eval current $\theta$ on all held-out examples}}
    \ENDFOR
    \ENDFOR

\end{algorithmic}
\end{algorithm}

\section{Results} \label{sec:results}


We evaluate two significantly different datasets, both in the few-shot regime.

\textbf{Omniglot}~(\citet{lake2015human}; handwritten characters, 1600 classes, 20 images per class) is a popular dataset for few-shot learning. Its large number of classes allows us to create very long, challenging trajectories for testing catastrophic forgetting under continual learning. However, due to its simple imagery we also include a dataset consisting of more complex natural images.

\textbf{Mini-ImageNet} (\citet{vinyals2016matching}; natural images, 100 classes, 600 images per class) contains hundreds of images per class, but in transfer we limit ourselves to 30 training images per class. This allows us to test the common scenario where we are allowed a large, diverse dataset in our pre-training, but our transfer-to dataset is of the limited few-shot variety.


For each pre-training configuration (Meta-ASB / ASB / i.i.d.\ and with/without {\zapping}; Section~\ref{sec:pre-training}), we report the average performance across 30 trials for transfer/continual learning results (3 random pre-train seeds and 10 random transfer seeds).  We sweep over three pre-training learning rates and seven transfer learning rates, and present the performance of the top performing learning rate for each configuration.
We only evaluate the models at the end of training (i.e.\ no early stopping), but the number of epochs is tuned separately for each training method and dataset so as to avoid overfitting. See Appendix~\ref{si:hyperparams} for more details on pre-training and hyperparameters.

Here we review the results on sequential transfer (Sec.~\ref{sec:olft}) and i.i.d.\ transfer (Sec.~\ref{sec:iid-transfer}). In the Appendix~\ref{sec:seq-transfer} we also show improvements due to {\zapping} in the \emph{unfrozen} sequential transfer setting. Furthermore, Appendix~\ref{si:zap-in-iid} shows more comparisons of i.i.d.\ pre-training with different amounts of {\zapping}, showing how {\zapping} alone can improve training but not as much as when it is combined with our ASB method.

\subsection{Continual Learning} \label{sec:olft}

\begin{figure}[b]
    \centering
    \includegraphics[width=\linewidth]{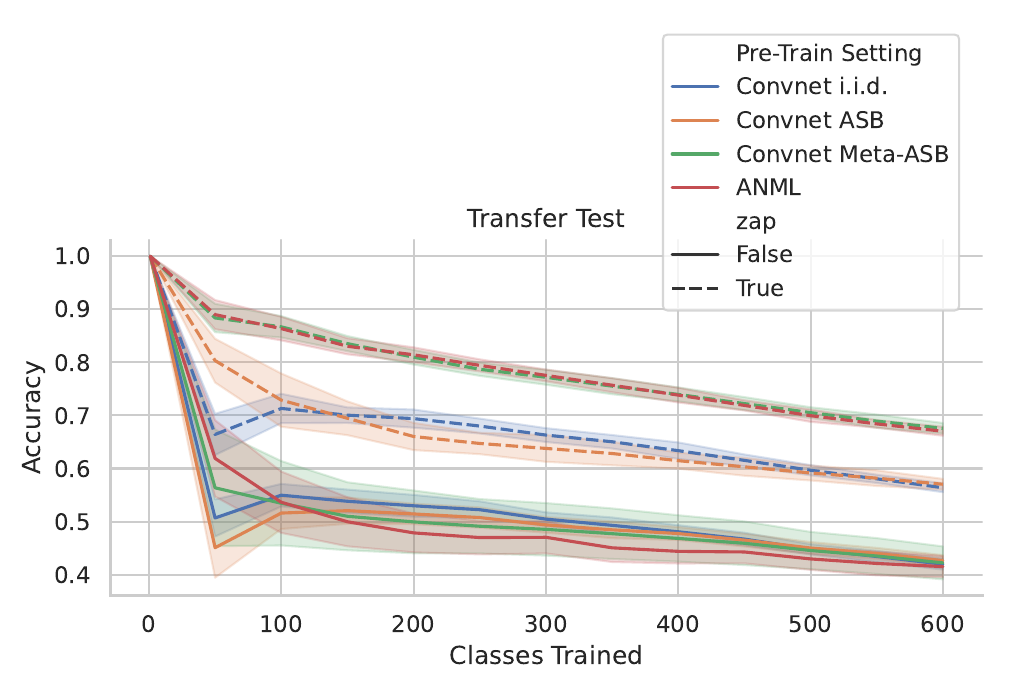}
    \caption{Sequential learning trajectories on Omniglot. Removing the neuromodulation layers from {\anml} has no impact on performance (Meta-ASB and {\anml} both achieve 67\% final accuracy). Removing {\zapping}, however, drastically affects performance, even when employing meta-learning. We do not compare directly to {\oml} since {\anml} represents the state of the art.}
    \label{fig:olft}
\end{figure}

\begin{figure*}[!htb]
    \centering
    \includegraphics[width=\linewidth]{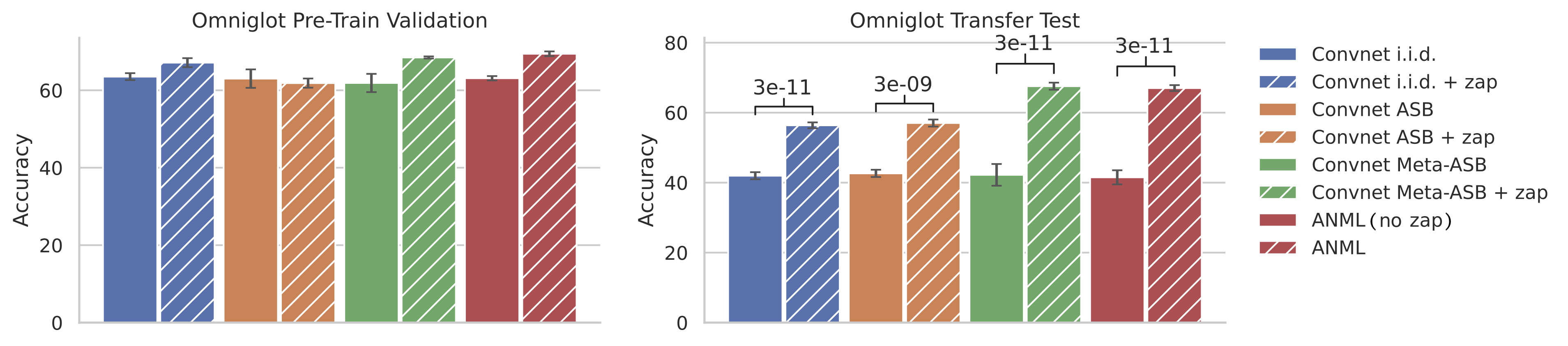} \\
    \vspace{1em}
    \caption{Average accuracy (and std dev error bars) for the sequential transfer learning problem, on Omniglot.
    \textbf{Pre-Train} is the final validation accuracy of the model on the \textit{pre-training} dataset. All the layers are trained during pre-train.
    \textbf{Transfer} is the accuracy on held-out instances from the \textit{transfer-to} dataset at the very end of sequential fine-tuning. Only the last layer is trained (linear probing) during transfer.
    Models trained \textbf{with {\zapping}} produce significantly ($p < 10^{-8}$) better transfer accuracy than their counterparts without {\zapping} in all cases (p-values of a two-sided Mann-Whitney U test are shown above each pair of bars). Note that the ANML model contains zapping by default and is therefore shaded in the legend.
    }
    \vspace{2em}
    \label{fig:olft-bars}
\end{figure*}

We evaluate the pre-trained variants described in~\ref{sec:pre-training} on the sequential transfer task as described in Section~\ref{sec:evaluation}. To quickly recap, the models are fine-tuned using linear probing on a few examples from classes not seen during pre-training, the examples are shown one at a time, and an optimization step is taken after each one.

As we see in Figure~\ref{fig:olft}, our Convnet Meta-ASB model performs similarly to the prior state of the art, {\anml}. In the prior work, {\anml} achieved 63.8\% accuracy after sequential learning of 600 classes. Our reimplementation shows a slightly higher performance of 67\% for both {\anml} and Meta-ASB.

However, our setup doesn't use heuristics on where/when in the model to apply optimization (like {\oml}; \citet{javed2019meta}), nor context-dependent gating (like {\anml}; \citet{beaulieu2020learning}), and uses fewer parameters than both prior works (see Appendix~\ref{si:models}).
This begs the question: 
what is it about these meta-learning algorithms that is contributing such drastic improvements? 

As an answer, we see from the solid green and red lines in Figure~\ref{fig:olft}, the models trained without {\zapping} show significantly lower performance (41.5\% and 42.2\% vs 67\%)---even though they \textit{were} trained with meta-learning. Figure~\ref{fig:olft-bars} shows that in all datasets, the meta-learned models with {\zapping} significantly outperform their non-{\zapping} counterparts, and outperform i.i.d.\ pre-training by an additional margin.

On Omniglot, the best model without {\zapping} achieves only 42.6\%. In fact, when applying {\zapping} to i.i.d.\ pre-training, we can even achieve better performance (56.4\%) than the models which are \textit{meta-learned} but \textit{without} {\zapping} (\mytilde42\%).  This suggests that {\zapping} may be an efficient and effective alternative (or complement) to meta-learning during pre-training.

\begin{figure}[b]
    \centering
    \includegraphics[width=\linewidth]{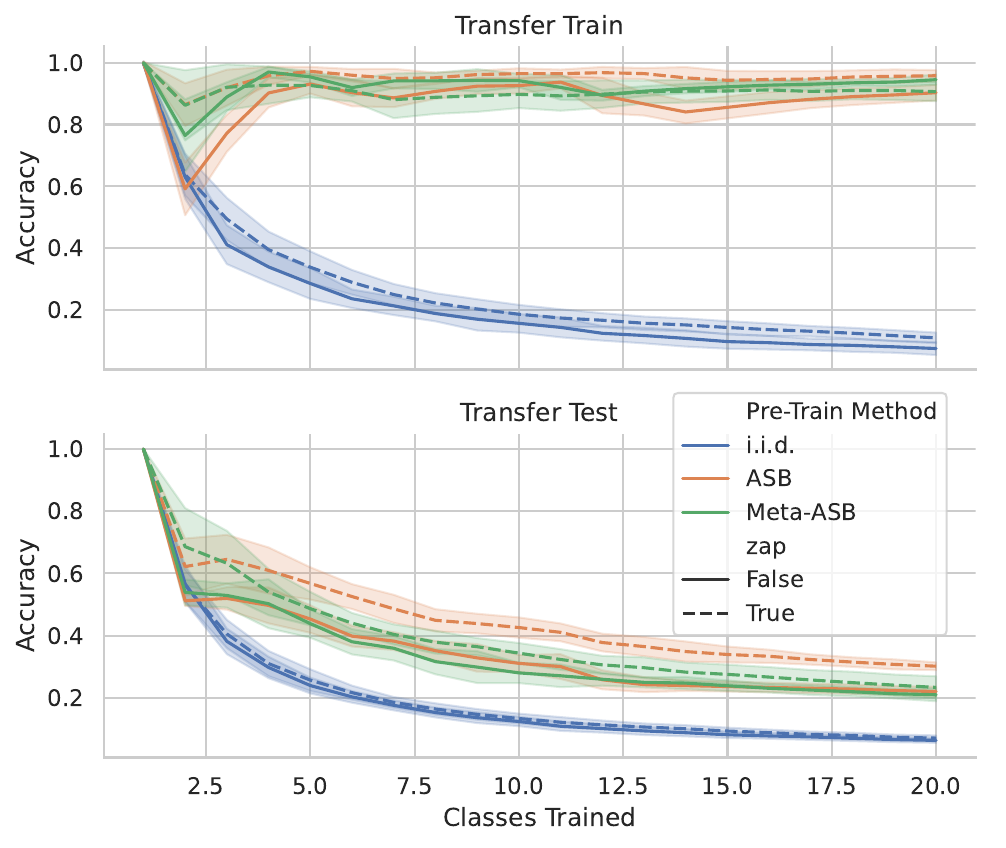}
    \caption{Accuracy on classes seen so far during continual transfer learning on Mini-ImageNet. 
    Models are trained on 30 examples from 20 new classes not seen during pre-training. All 30 images from a class are shown sequentially one at a time before switching to the next class. 
    After each class, validation accuracy on the transfer set is measured using 100 examples per class, from all classes seen up to that point.
    Models pre-trained \textbf{with ASB} (with or without meta-gradients) significantly outperform i.i.d. pre-training.
    ASB+{\zapping} further outperforms plain ASB ($p < 10^{-10}$).}
    \label{fig:mini-imagenet-olft}
\end{figure}

On Mini-ImageNet (Figure~\ref{fig:mini-imagenet-olft}), we again see a substantial difference between {\zapping} models and their non-{\zapping} counterparts (except for i.i.d.+{\zap}).
While on Omniglot Meta-ASB+{\zap} outperformed ASB+{\zap}, on Mini-ImageNet we observe that ASB+{\zap} achieves the best accuracy, further demonstrating the effectiveness of the ASB method as a way to emulate the challenges of transfer during pre-training.

In Figure~\ref{fig:olft-bars}, we also include the \textit{pre-train validation accuracy}: this is the validation accuracy of the pre-trained model \textit{on the pre-training dataset}, before it was modified for transfer. We observe that ranking models by validation performance \textit{is not well correlated with ranking of transfer performance}.
This lack of pre-train $\leftrightarrow$ transfer correlation introduces a dilemma, whereby we may not have a reliable way of judging which models will be better for transfer until we actually try them.

Across all three datasets, we have observed that:
\begin{enumerate}
    \item {\Zapping} is a significant driver of performance improvements (see dashed vs.\ solid lines per treatment in Figures~\ref{fig:olft} and~\ref{fig:mini-imagenet-olft}).
    \item {\Zapping} sometimes also improves normal \mbox{(pre-)training} performance, although this trend is less consistent (Figure~\ref{fig:olft-bars}; more details in Appendix~\ref{si:full-results}, Table~\ref{tab:olft}).
    \item Counter-intuitively, even the  Alternating Sequential and Batch learning (ASB) sampling routine alone (without meta-gradients) appears to provide some benefits for transfer (see \textcolor{C1}{ASB} vs \textcolor{C0}{i.i.d.}\ in Figure \ref{fig:mini-imagenet-olft}). It may sometimes be unnecessary to use the much more expensive and complex higher-order gradients.
\end{enumerate}

\subsection{Transfer Learning} \label{sec:iid-transfer}

\begin{figure}[!htb]
    \centering
    \subfigure[Omniglot (15 training images / 5 testing images per class).]{
        \includegraphics[width=\linewidth]{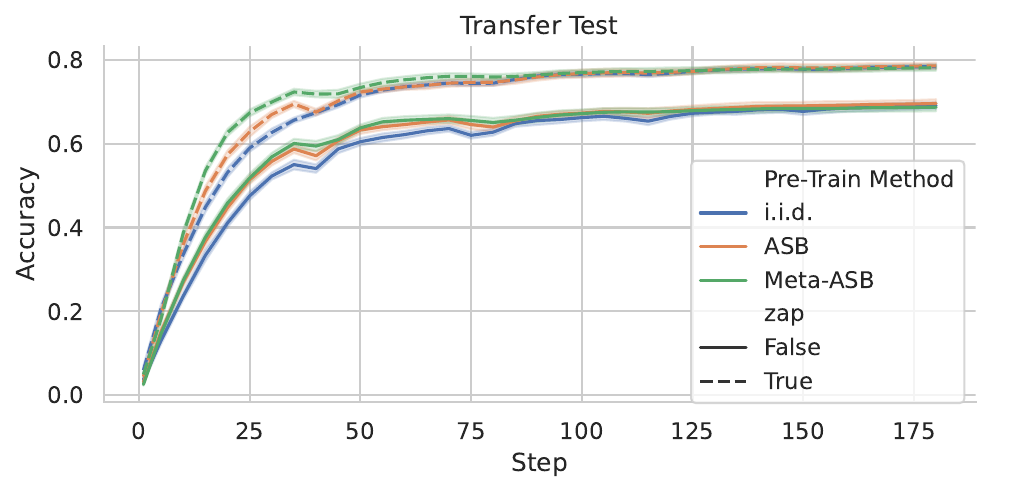}
        \label{fig:omniglot-iid}
    }
    \hfill
    \vspace{1em}
    \subfigure[Mini-ImageNet (30 training / 100 testing images per class).]{
        \includegraphics[width=\linewidth]{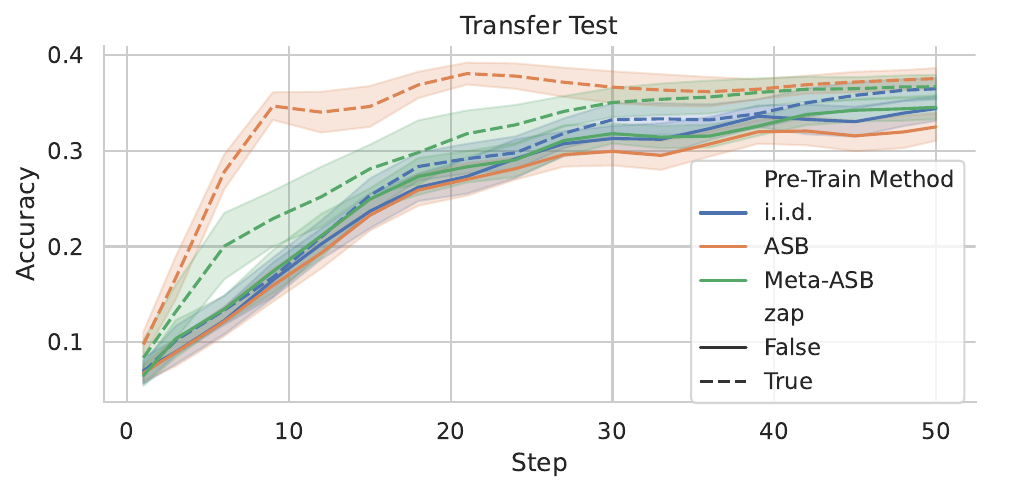}
        \label{fig:mini-imagenet-iid}
        \vspace{1em}
    }
    \vspace{1em}
    \caption{Validation accuracy over training time on all classes in the transfer set during fine-tuning with standard i.i.d.\ batches.  For 
    all datasets, 
    models pre-trained \textbf{with {\zapping}} achieve significantly higher transfer accuracy at end of fine-tuning. 
    While ASB methods (green, orange) do not dramatically improve final performance, they achieve more rapid fine-tuning relative to i.i.d.+zap pre-training (blue).
    }
    \vspace{2em}
    \label{fig:iid-transfer}
\end{figure}

Although the zapping and meta-learning methods described in Algorithm \ref{alg:pretraining} were originally designed to learn robust representations for continual learning, we show that they are beneficial for transfer learning in general. Here we feature the results of standard i.i.d.\ transfer, as described in Section~\ref{sec:evaluation}. We train each model for five epochs of unfrozen fine-tuning on mini-batches of data using the Adam optimizer.  

Figure~\ref{fig:iid-transfer} shows results on Omniglot and Mini-ImageNet. As in the continual learning tests, here we also see substantial gains for the models employing {\zapping} over those that do not.  When {\zapping} is not employed, models pre-trained with meta-gradients are comparable to those trained simply with standard i.i.d.\ pre-training. See Table~\ref{tab:iid} in Appendix~\ref{si:full-results} for a detailed comparison of final values.

Despite both the zapping and ASB pre-training methods stemming from attempts to reduce catastrophic forgetting in continual learning settings, zapping consistently provides advantages over non-zapped models for all pre-training configurations on standard i.i.d.\ transfer learning. 
We hypothesize that these two settings---continual and transfer learning---share key characteristics that make this possible. Both cases may benefit from an algorithm which produces more adaptable, robust features that can quickly learn new information while preserving prior patterns that may help in future tasks.

\subsection{Toward Larger Architectures}
\label{sec:bigger_is_better}

We conclude our investigation showing how {\zapping} and ASB influence the training dynamics of a larger model, specifically the widely-used VGG-16 architecture \citep{simonyan2014very}. For this larger model, the simplicity of Omniglot images presents a limitation; therefore we use a variant called Omni-image \citep{omnimage23}. The Omni-image dataset retains the task structure of Omniglot (i.e. 20 images per class, high within-class visual consistency) but uses natural images (instead of handwritten characters) taken from the 1000 classes available in ImageNet-1k \citep{deng2009imagenet}. Omni-image was designed for few-shot and continual learning, and we show some examples in Appendix \ref{si:omnimage}, Figure \ref{SI:mini_omni_image}.

\begin{figure}[!ht]
    \centering
    \includegraphics[width=1\linewidth]{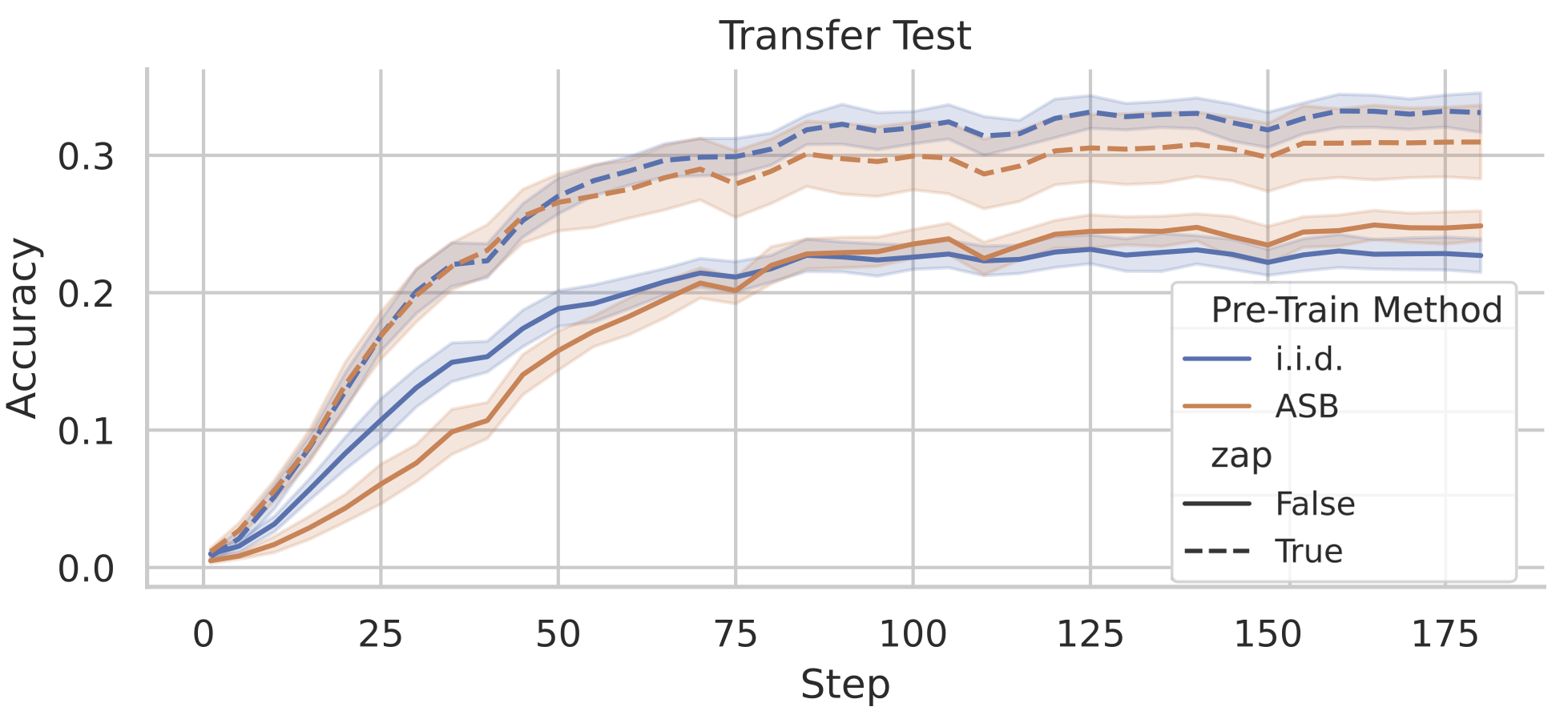}
    \vspace{0.5em}
    \caption{VGG-16 transfer on the Omni-image dataset. {\Zapping} improves both the speed of adaptation and the final accuracy, but ASB does not contribute any further improvement.}
    \vspace{2em}
    \label{fig:vgg}
\end{figure}

Figure \ref{fig:vgg} shows that models trained using zapping not only learn faster but also achieve a better final performance. These results suggest that zapping and ASB could potentially be applied to other architectures but the advantages of ASB may depend on the specific task and dataset structure. For instance, consider the differing results observed with Omniglot and Omni-image (Figures \ref{fig:olft-bars} \& \ref{fig:vgg}). ASB+{\zap} outperformed i.i.d.+{\zap} in the former, but was slightly worse in the latter.

\section{Discussion}

Across three substantially different datasets, {\zapping} consistently results in better representations for transferring to few-shot datasets, leading to improvements in both a continual learning and standard transfer setting. 
In many cases, we are still able to achieve the best performance by just applying {\zapping} and alternating optimizations of sequential learning and batch learning (ASB), without applying any meta-gradients.

We even see some benefit from applying {\zapping} directly to i.i.d.\ training, without any sequential learning component. 
This raises the question of whether we can match the performance of meta-learning using only {\zapping} and standard i.i.d.\ training. However, this setting introduces new choices of when and where to reset neurons, since we are learning in batches and not just one class at a time. 
We include ablations in Appendix~\ref{si:zap-in-iid} that examine this question; in most cases, more {\zapping} leads to better performance, but it is still outmatched by ASB.
This serves as a promising starting point, and better variants of this scheme could likely be discovered. 

It is reasonable to suppose that the constant injection of noise by resetting weights during training helps to discover weights which are not as affected by this disruption, thus building some resilience to the shock of re-initializing layers.
If it can be shown that the noise injections reduce the \textit{co-adaptation of layers} \citep{yosinski2014transferable}, thereby increasing their resilience, it begs the question of how it relates to other co-adaptation reducing mechanisms such as dropout, which is also shown to improve continual \citep{mirzadeh2020dropout} and transfer \citep{semwal2018practitioners} learning.

The approaches explored here include pre-training by alternating between sequential learning on a single class and batches sampled from all pre-training classes (ASB), and resetting classifier weights prior to training on a new class ({\zapping}). The information accumulated by repeating these simple methods across many tasks during the pre-training process mimics the condition experienced during transfer learning at test time. Episodes of sequential training are likely to cause both catastrophic forgetting and overfitting. Models that manage to overcome those hurdles during training seem to develop resilient features, beyond what the loss function selects for. We thus argue that the results above demonstrate a simple yet effective alternative to meta-learning---one without expensive meta-gradients to backpropagate through tasks.  

\section{Related work}

As we have shown, the {\zapping} operation of resetting last-layer weights provides clear performance improvements, but what is it about continually injecting random weights that enables this improved learning?

The work of \citet{frankle2020early} investigates the dynamics of learning during the early stages of network training. They show that large gradients lead to substantial weight changes in the beginning of training, after which gradients and weight signs stabilize.
Their work suggests that these initial drastic changes are linked to weight-level noise. 

\citet{dohare2021continual} also investigate the relationship between noise and learning, showing that stochastic gradient descent is not enough to learn continually, and that this problem can be alleviated by repeated injections of noise.
Rather than resetting classification neurons of the last layer, they choose weights to reset based on a pruning heuristic.

The reinitialization of weights in a neural network during training is an interesting emerging topic, with many other works investigating this phenomenon in a number of different settings.
Like us, \citet{zhao2018retraining} periodically reinitialize the last layer of a neural network during training. Their focus is on ensemble learning for a single dataset, rather than transfer learning. 
%
\citet{li2020rifle} also periodically reinitialize the last layer, but they do it \textit{during transfer}, rather than pre-training, which may not be possible depending on the application. 
%
Both \citet{alabdulmohsin2021impact} and \citet{zhou2022fortuitous} investigated the idea of reinitialization of upper layers of the network, building upon the work of~ \citet{taha2021knowledge}. They show performance improvements in the few-shot regime. However, they focus on learning of a single dataset rather than transfer learning. 
The same is true of \citet{zaidi2022does}, who evaluate an extensive number of models to find under which circumstances reinitialization helps. 

\citet{nikishin2022primacy} apply a similar mechanism to deep reinforcement learning. They find that periodically resetting the final layers of the Q-value network is beneficial across a broad array of settings. 
\citet{ramkumar2023learn} study the application of resetting to a version of online learning where data arrives in mega-batches. They employ resetting as a compromise between fine-tuning a pre-trained network and training a new network from scratch. 
More generally, \citet{lyle2023understanding} attempt to discover the reasons for plasticity loss in neural networks, and show the resetting of final layers to be one of a few effective methods in maintaining plasticity.

One unique aspect of our work is the {\zapping} + ASB routine, where we forget one class at a time and focus on relearning that class. Another major difference from prior investigations is that we examine how resetting weights better prepares a pre-trained model for transfer learning. In this regard, the concurrent work of \citet{chen2023improving} examines the same topic, albeit in the domain of natural language processing. Their method repeatedly forgets the early embedding layers of a language model, rather than the later classification layers of an image model. As with our work, they also find that this repeated forgetting results in a meta-learning-like effect which better prepares the model for downstream transfer. This gives us a new lens through which to view weight resetting.

\section{Conclusion \& Future Work}

We have revealed the importance of ``{\zapping}'' and relearning for pre-training, and its connection to meta-learning. We have shown that {\zapping} can lead to significant gains in transfer learning performance across multiple settings and datasets. The concept of forgetting and relearning has been investigated in other recent works, and our observations add to the growing evidence of the usefulness of this concept.

Aside from the benefits of {\zapping} and relearning, our results highlight the disruptive effect of the re-initialization of last layers in general. Resetting of the final layer is routine in the process of fine-tuning a pre-trained model \citep{yosinski2014transferable}, but the impact of this ``transfer shock'' is still not fully clear. For instance, it was only recently observed that fine-tuning in this way underperforms on out-of-distribution examples, and \citet{kumar2022finetuning} suggest to freeze the lower layers of a network to allow the final layer to stabilize. 
A deeper understanding of these mechanisms could significantly benefit many areas of neural network research.

Finally, this work explores a simpler approach to meta-learning than meta-gradient approaches. It does so by repeatedly creating transfer shocks during pre-training, encouraging a network to learn to adapt to them.  Future work should explore other methods by which we can approximate transfer learning during pre-training, how to influence the features learned to maximize transfer performance with the most computational efficiency, and how the benefits of {\zapping} scale to larger models.




\begin{ack}
We would like to thank Sara Pelivani for pointing out that the neuromodulatory network in ANML was not necessary, and for the interesting discussion that resulted from this observation. This material is based upon work supported by the Broad Agency Announcement Program and Cold Regions Research and Engineering Laboratory (ERDCCRREL) under Contract No. W913E521C0003, National Science Foundation under Grant No. 2218063, and Defense Advanced Research Projects Agency under Cooperative Agreement No. HR0011-18-2-0018. Computations were performed on the Vermont Advanced Computing Core supported in part by NSF Award No. OAC-1827314, and also by hardware donations from AMD as part of their HPC Fund.
\end{ack}



\bibliography{references}

\newpage

\appendix
\onecolumn

\section{Full Result Tables} \label{si:full-results}

Here we include the exact numbers corresponding to the plots shown in the main text of the paper, for all treatments on all datasets.

\begin{table}[htbp]
    \begin{center}
    \begin{tabular}{cc@{\hskip 0.2in}rr@{\hskip 0.2in}rr@{\hskip 0.2in}rr}

    \toprule
    \multicolumn{1}{c}{\bf Pre-Train Method}
    &\multicolumn{1}{c@{\hskip 0.2in}}{\bf Zap}
    &\multicolumn{2}{c@{\hskip 0.2in}}{\bf Omniglot}
    &\multicolumn{2}{c@{\hskip 0.2in}}{\bf Mini-ImageNet} \\
    & 
    &\multicolumn{1}{c}{\bf Pre-Train}  &\multicolumn{1}{c@{\hskip 0.2in}}{\bf Transfer}
    &\multicolumn{1}{c}{\bf Pre-Train}  &\multicolumn{1}{c@{\hskip 0.2in}}{\bf Transfer}\\
    \midrule

    i.i.d.\      &              & 63.5 {\scriptsize $\pm 0.9$} & 42.0 {\scriptsize $\pm 1.0$} & 45.4 {\scriptsize $\pm 1.3$} &  6.3 {\scriptsize $\pm 0.8$} \\
    i.i.d.\      & \greencheck  & 67.1 {\scriptsize $\pm 1.2$} & 56.4 {\scriptsize $\pm 0.8$} & 47.5 {\scriptsize $\pm 0.9$} &  7.3 {\scriptsize $\pm 0.8$} \\
    ASB          &              & 63.0 {\scriptsize $\pm 2.4$} & 42.6 {\scriptsize $\pm 1.0$} & 31.0 {\scriptsize $\pm 0.9$} & 22.0 {\scriptsize $\pm 1.5$} \\
    ASB          & \greencheck  & 61.9 {\scriptsize $\pm 1.2$} & 57.0 {\scriptsize $\pm 1.0$} & 36.3 {\scriptsize $\pm 1.9$} & \textbf{30.2} {\scriptsize $\pm 1.2$} \\
    Meta-ASB     &              & 61.9 {\scriptsize $\pm 2.4$} & 42.2 {\scriptsize $\pm 3.1$} & 30.4 {\scriptsize $\pm 0.6$} & 21.0 {\scriptsize $\pm 2.1$} \\
    Meta-ASB     & \greencheck  & 68.5 {\scriptsize $\pm 0.3$} & \textbf{67.6} {\scriptsize $\pm 1.0$} & 17.1 {\scriptsize $\pm 7.2$} & 23.3 {\scriptsize $\pm 3.6$} \\
    ANML w/o zap &              & 63.1 {\scriptsize $\pm 0.6$} & 41.5 {\scriptsize $\pm 2.0$} & -  & - \\
    ANML         & \greencheck  & 69.4 {\scriptsize $\pm 0.6$} & \textbf{67.0} {\scriptsize $\pm 0.8$} & -  & - \\

    \bottomrule
    \end{tabular}
    \end{center}
    \caption{Average accuracy ($\pm$ std dev) of the best model in each category, for the sequential transfer problem.
    \textbf{Pre-Train} is the final validation accuracy of the model on the \textit{pre-training} dataset.
    \textbf{Transfer} is the accuracy on held-out instances from the \textit{transfer-to} dataset at the very end of sequential fine-tuning (i.e.\ after training on all classes).
    The best transfer configuration for each dataset (and those within one std dev) are highlighted in bold.
    Models trained \textbf{with {\zapping}} result in significantly better transfer accuracy than those trained without {\zapping} for all pre-training methods and datasets (when comparing the distribution of Zap~{\greencheck} accuracies to their non-zapped counterparts under a two-sided Mann-Whitney U test, all p-values are under 1e-8).
    }
    \label{tab:olft}
\end{table}

\begin{table}[htbp]
    \begin{center}
    \begin{tabular}{cc@{\hskip 0.2in}rr@{\hskip 0.2in}rr@{\hskip 0.2in}rr}

    \toprule
    \multicolumn{1}{c}{\bf Pre-Train Method}
    &\multicolumn{1}{c@{\hskip 0.2in}}{\bf Zap}
    &\multicolumn{2}{c@{\hskip 0.2in}}{\bf Omniglot}
    &\multicolumn{2}{c@{\hskip 0.2in}}{\bf Mini-ImageNet} \\
    & 
    &\multicolumn{1}{c}{\bf Pre-Train}  &\multicolumn{1}{c@{\hskip 0.2in}}{\bf Transfer}
    &\multicolumn{1}{c}{\bf Pre-Train}  &\multicolumn{1}{c@{\hskip 0.2in}}{\bf Transfer}\\
    \midrule

    i.i.d.\    &      & 63.5 {\scriptsize $\pm 0.9$} & 69.0 {\scriptsize $\pm 0.8$} & 45.4 {\scriptsize $\pm 1.3$} & 34.4 {\scriptsize $\pm 1.3$}\\
    i.i.d.\    & \greencheck    & 67.1 {\scriptsize $\pm 1.2$} & \textbf{78.4} {\scriptsize $\pm 0.5$} & 47.5 {\scriptsize $\pm 0.9$} & \textbf{36.5} {\scriptsize $\pm 1.1$}\\
    ASB        &     & 63.0 {\scriptsize $\pm 2.4$} & 69.6 {\scriptsize $\pm 1.0$} & 43.9 {\scriptsize $\pm 2.0$} & 32.5 {\scriptsize $\pm 1.4$}\\
    ASB        & \greencheck    & 61.9 {\scriptsize $\pm 1.2$} & \textbf{78.5} {\scriptsize $\pm 0.8$} & 36.3 {\scriptsize $\pm 1.9$} & \textbf{37.6} {\scriptsize $\pm 1.1$} \\
    Meta-ASB   &     & 63.5 {\scriptsize $\pm 1.5$} & 68.7 {\scriptsize $\pm 1.0$} & 45.1 {\scriptsize $\pm 1.3$} & 34.6 {\scriptsize $\pm 1.2$}\\
    Meta-ASB   & \greencheck    & 68.2 {\scriptsize $\pm 0.9$} & \textbf{77.8} {\scriptsize $\pm 0.9$} & 17.1 {\scriptsize $\pm 7.2$} & \textbf{36.7} {\scriptsize $\pm 1.2$}\\
    ANML w/o zap   &     & 65.8 {\scriptsize $\pm 0.6$} & 70.4 {\scriptsize $\pm 0.8$} & - & -\\
    ANML       & \greencheck    & 69.4 {\scriptsize $\pm 0.6$} & \textbf{78.2} {\scriptsize $\pm 0.7$} & - & -\\

    \bottomrule
    \end{tabular}
    \end{center}
    \caption{Average accuracy ($\pm$ std dev) for the standard fine-tuning transfer problem. 
    \textbf{Pre-Train} is the final validation accuracy of the model on the \textit{pre-training} dataset.
    \textbf{Transfer} is the accuracy on held-out instances from the \textit{transfer-to} dataset after five epochs of fine-tuning.
    The best transfer configuration for each dataset (and those within one std dev) are highlighted in bold.
    Models trained \textbf{with {\zapping}} produce significantly better transfer accuracy than those trained without {\zapping} for all pre-training methods and datasets (when comparing the distribution of Zap~{\greencheck} accuracies to their non-zapped counterparts under a two-sided Mann-Whitney U test, all p-values are under 1e-6).
    }
    \label{tab:iid}
\end{table}

\begin{table}[htbp]
    \begin{center}
    \begin{tabular}{rc@{\hskip 0.2in}rr@{\hskip 0.2in}rr@{\hskip 0.2in}rr}

    \toprule
    \multicolumn{1}{c}{\bf Pre-Train Method}
    &\multicolumn{1}{c@{\hskip 0.2in}}{\bf Zap}
    &\multicolumn{2}{c@{\hskip 0.2in}}{\bf Mini-ImageNet}
    &\multicolumn{2}{c@{\hskip 0.2in}}{\bf Omni-Image} \\
    & 
    &\multicolumn{1}{c}{\bf Pre-Train}  &\multicolumn{1}{c@{\hskip 0.2in}}{\bf Transfer}
    &\multicolumn{1}{c}{\bf Pre-Train}  &\multicolumn{1}{c@{\hskip 0.2in}}{\bf Transfer}\\
    \midrule

    i.i.d.\      &              & 51.6 {\scriptsize $\pm 1.2$} & \textbf{46.2} {\scriptsize $\pm 1.6$} & 27.5 {\scriptsize $\pm 3.2$} &         22.7  {\scriptsize $\pm 1.2$} \\
    i.i.d.\      & \greencheck  & 53.3 {\scriptsize $\pm 0.8$} & \textbf{47.5} {\scriptsize $\pm 1.6$} & 25.2 {\scriptsize $\pm 2.2$} & \textbf{33.1} {\scriptsize $\pm 1.4$} \\
    ASB          &              & 43.2 {\scriptsize $\pm 3.5$} &         44.5  {\scriptsize $\pm 1.7$} & 29.1 {\scriptsize $\pm 3.7$} &         24.9  {\scriptsize $\pm 1.1$} \\
    ASB          & \greencheck  & 48.2 {\scriptsize $\pm 3.5$} & \textbf{46.4} {\scriptsize $\pm 1.2$} & 25.9 {\scriptsize $\pm 0.8$} & \textbf{31.0} {\scriptsize $\pm 2.7$} \\

    \bottomrule
    \end{tabular}
    \end{center}
    \caption{Average accuracy ($\pm$ std dev) of the best model in each category, for standard transfer on \textbf{VGG-16}.
    \textbf{Pre-Train} is the final validation accuracy of the model on the \textit{pre-training} dataset.
    \textbf{Transfer} is the accuracy on held-out instances from the \textit{transfer-to} dataset after five epochs of fine-tuning.
    The best transfer configuration for each dataset (and those within one std dev) are highlighted in bold.
    Models trained \textbf{with {\zapping}} result in significantly better transfer accuracy than those trained without {\zapping} for all pre-training methods and datasets (when comparing the distribution of Zap~{\greencheck} accuracies to their non-zapped counterparts under a two-sided Mann-Whitney U test, all p-values are under 1e-8).
    }
    \label{tab:vgg}
\end{table}

\section{Separate Weights for Inner and Outer Loops}
\label{si:models}

\begin{figure}[htb!]
  \centering
  \subfigure[{\oml} architecture]{
      \includegraphics[width=0.31\linewidth]{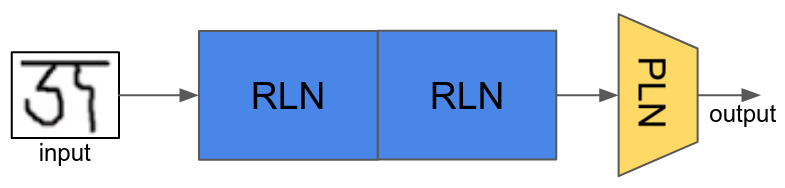}
      \label{fig:oml_arch}
  }
  \hfill
  \subfigure[{\anml} architecture]{
      \includegraphics[width=0.31\linewidth]{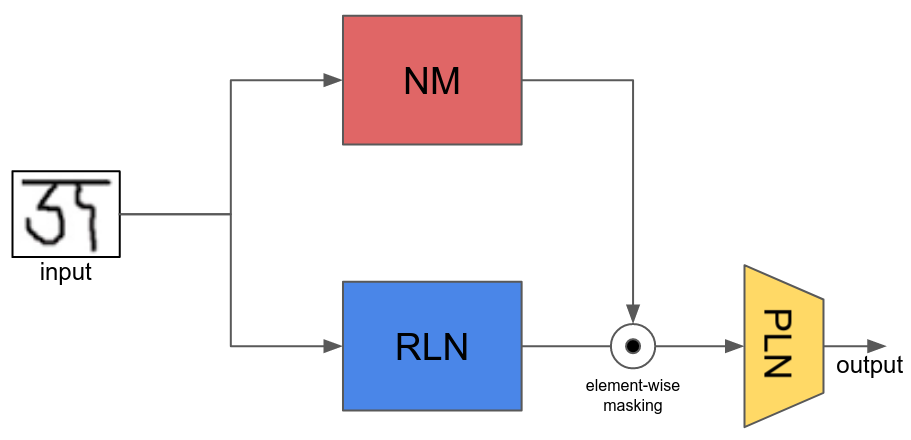}
      \label{fig:anml_arch}
  }
  \hfill
  \subfigure[{\sanml} architecture]{
      \includegraphics[width=0.31\linewidth]{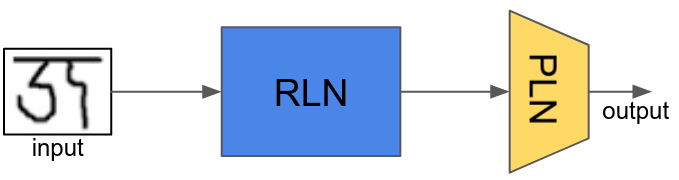}
      \label{fig:sanml_arch}
  }
  \vspace{1em}
  \caption{{\oml}, {\anml} and, {\sanml} architectures}
  \vspace{2.5em}
  \label{fig:oml_anml_arch}
\end{figure}

As mentioned in Section \ref{sec:methods}, the meta-train phase is split between inner and outer loops. 
To incentivize the discovery of generalizable features the {\anml} and {\oml} models train different parts of the model at different times. The last layer weights ({\pln} in Figure \ref{fig:oml_arch}) are only updated in the inner loop while the whole network ({\rln} + {\pln} in Figure \ref{fig:oml_arch}) are updated only in the outer loop. The rationale of this choice is that the network can leverage meta-gradients in the outer loop to find features that can improve the inner loop.
In a similar fashion the {\anml} model only updates the neuromodulation weights ({\nm} in Figure \ref{fig:anml_arch}) using meta-gradients in the outer loop, while instead the rest of the weights ({\rln} + {\pln} in Figure \ref{fig:anml_arch}). This design choice aims to break the symmetry of inner/outer loops and incentivize the outer loop to refine/integrate what was learned during the inner loop.
Unlike ANML and OML, our {\sanml} updates all weights in every phase. We remove the NM layers from the {\anml} model, yet perform just as well in one-layer fine-tuning (Figure~\ref{fig:olft}).
Our architecture is also shallower than {\oml} (3 conv layers instead of 6).

\begin{table}[ht!]
\centering
\begin{tabular}{l|c|c|c||c|c|c}
      & \multicolumn{3}{c||}{Inner (SGD)} & \multicolumn{3}{c}{Outer (meta-Adam)} \\ \cline{2-7} 
      & NM     & RLN     & PLN     & NM     & RLN     & PLN    \\ \hline
{\oml}  & -      & \redcross       & \greencheck       & -      & \greencheck       & \greencheck      \\
{\anml}  & \redcross      & \greencheck       & \greencheck       & \greencheck      & \greencheck       & \greencheck      \\
{\sanml} & -      & \greencheck       & \greencheck       & -      & \greencheck       & \greencheck     
\end{tabular}
\vspace{1em}
\caption{\label{tab:phases} Different update strategies}
\vspace{1.5em}
\end{table}

\section{Hyperparameters}
\label{si:hyperparams}

We use the Adam optimizer \citep{kingma2014adam} with a standard cross-entropy loss. We train all models to convergence on all our datasets, and we take the final checkpoint as our pre-trained model for continual/transfer tests. In the i.i.d.\ pre-training setting, we do not have the single-class inner loop, so we must decide how often to zap in a different manner. We investigate the effect of {\zapping} at different frequencies and a different number of classes (from a single one to all of them at once) to determine the optimal configuration (Appendix~\ref{si:zap-in-iid}).

See Tables~\ref{tab:asb-hyperparams} and~\ref{tab:iid-hyperparams} for a listing of pre-training hyperparameters used for our experiments.

\begin{table}[ht!]
    \begin{center}
    \begin{tabular}{lrr}

    \toprule
    {\bf Parameter}
    &{\bf Omniglot }
    &{\bf Mini-ImageNet} \\
    \midrule

    training examples per class & 15 & 500 \\
    validation examples per class & 5 & 100 \\
    inner loop steps & 20 & same \\
    ``remember set'' size & 64 & 100 \\  
    inner optimizer & SGD & same \\
    inner learning rates & [0.1, 0.01, 0.001] & same \\
    outer optimizer & Adam & same \\
    outer learning rates & [0.1, 0.01, 0.001] & same \\
    outer loop steps & 9,000\tablefootnote{In Meta-ASB, we train for 25,000 steps instead of 9,000. It is not usually necessary to train beyond \mytilde 18,000 steps, but we do typically need to train longer than non-meta-ASB, and we don't usually see any detriment in training longer than needed.} & same \\

    \bottomrule
    \end{tabular}
    \end{center}
    \caption{Hyperparameters used for pre-training using Alternating Sequential and Batch Learning (ASB).}
    \label{tab:asb-hyperparams}
\end{table}

\begin{table}[ht!]
    \begin{center}
    \begin{tabular}{lrr}

    \toprule
    {\bf Parameter}
    &{\bf Omniglot }
    &{\bf Mini-ImageNet} \\
    \midrule

    training examples per class & 15 & 500 \\
    validation examples per class & 5 & 100 \\
    Adam learning rates & [3e-4, 1e-3, 3e-4] & same \\
    batch size & 256 & same \\
    epochs & 10 / 30 & 30 \\

    \bottomrule
    \end{tabular}
    \end{center}
    \caption{Hyperparameters used for pre-training using standard i.i.d.\ batch learning.}
    \label{tab:iid-hyperparams}
\end{table}

\clearpage
\section{Network Architecture}
\label{si:convnets}

\begin{figure}[ht]
    \centering
    \includegraphics[width=0.4\linewidth]{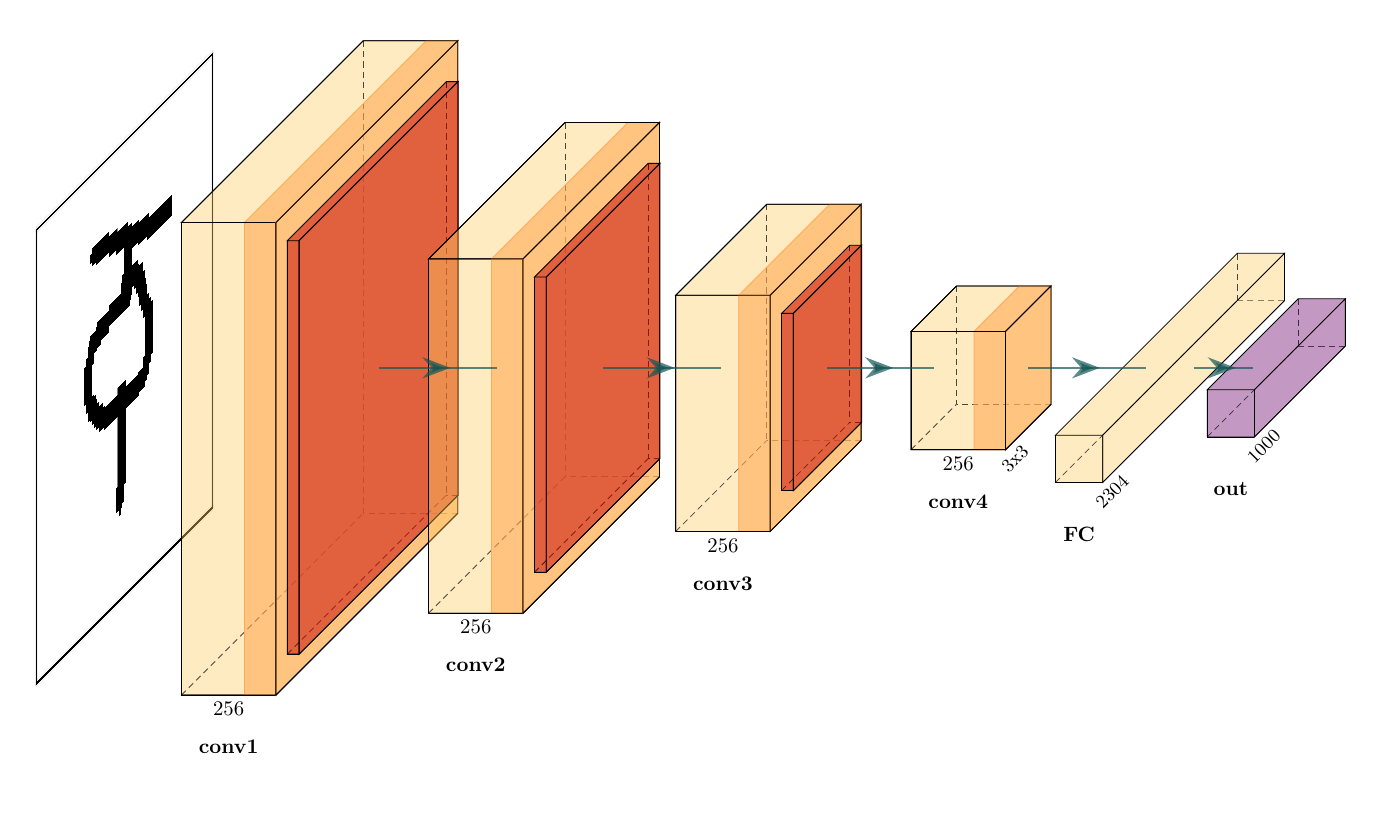}
    \vspace{1em}
    \caption{Convnet4, a standard architecture with four convolutional blocks and one fully-connected linear layer. The example in this case corresponds to a hypothetical dataset which has 1,000 target classes.}
    \vspace{2.5em}
    \label{fig:si-convnet}
\end{figure}

See Figure~\ref{fig:si-convnet} for a depiction of the neural network architecture used in this work. The Mini-ImageNet and OmnImage datasets consist of images of size 84x84. For these, we use a typical architecture consisting of four convolutional blocks and one fully-connected final layer. Each convolutional block consists of: convolution, InstanceNorm \citep{ulyanov2016instance}, ReLU activation, and max pool layers, in that order. All convolutional layers have 256 output channels. An architecture similar to this has been used to good effect for much exploratory research in few-shot learning---in particular, we were inspired by ``Few-Shot Meta-Baseline'' \citep{chen2021meta}.

For Omniglot, we use 28x28 single-channel images, and so the architecture is slightly different. Instead of four convolutional blocks, we use three. Also, we skip the final pooling layer.

\vfill
\vspace{1cm}

\section{Alternating Sequential and Batch Learning}
\label{si:alternating}

\begin{figure}[hb]
    \centering
    \subfigure[Meta-ASB]{
        \includegraphics[width=0.5\linewidth]{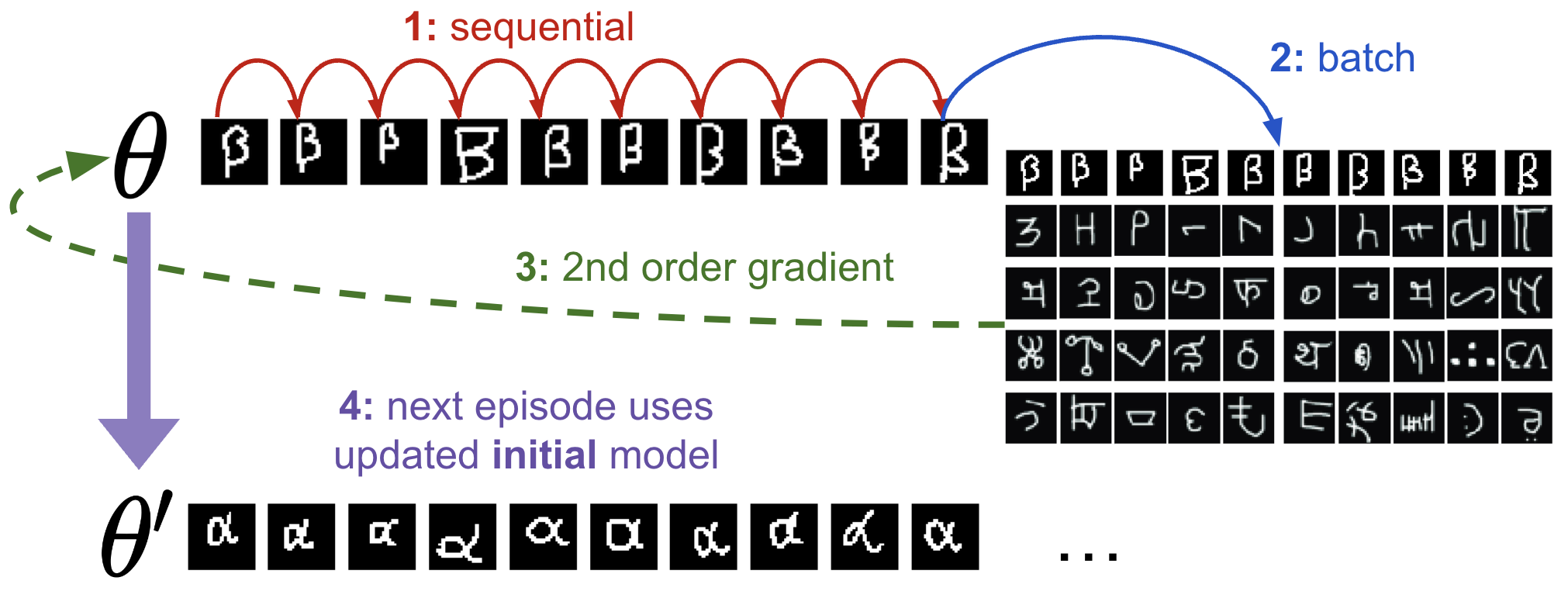}
        \label{si:alternating-meta}
    } \\
    \vspace{1em}
    \subfigure[ASB]{
        \includegraphics[width=0.5\linewidth]{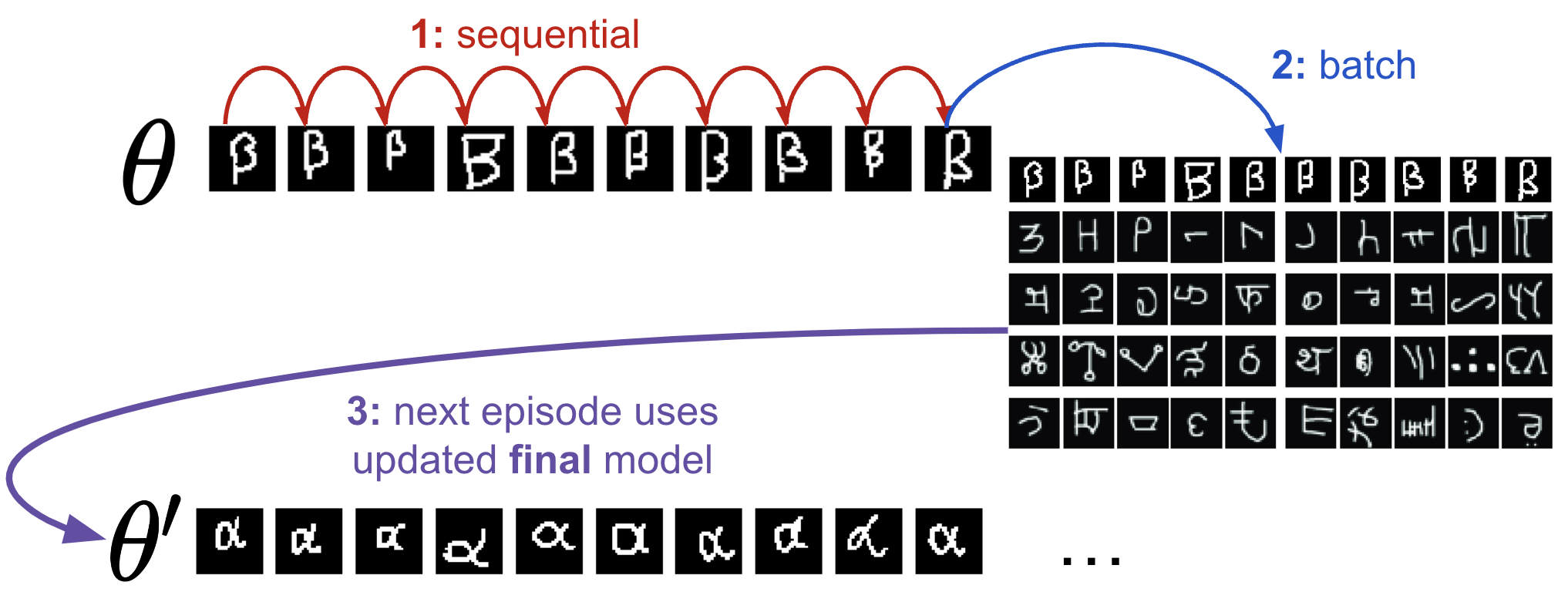}
        \label{si:alternating-nometa}
    } \\
    \vspace{1em}
    \caption{The ASB training procedure with and without higher-order gradients based meta-learning.}
    \vspace{2em}
    \label{fig:si-alternating}
\end{figure}

See Figure~\ref{fig:si-alternating} for a visual depiction of the Alternating Sequential and Batch (ASB) learning procedure, and its meta-learning variant.

\clearpage
\section{Performing {\Zapping} in i.i.d. Pre-Training}
\label{si:zap-in-iid}

During ASB, the model is presented with episodes of sequential learning on a single class, so we always reset the last layer weights corresponding to the class which is about to undergo sequential learning. In the i.i.d.\ case, there is no such "single task training phase". Instead, we choose to reset $K$ classes (out of $N$ total classes) on a cadence of once per $E$ epochs. This introduces two new hyperparameters $(K, E)$ for us to sweep over. 

\begin{figure}[ht]
    \centering
    \subfigure[Omniglot Sequential Transfer.]{
        \includegraphics[width=0.43\linewidth]{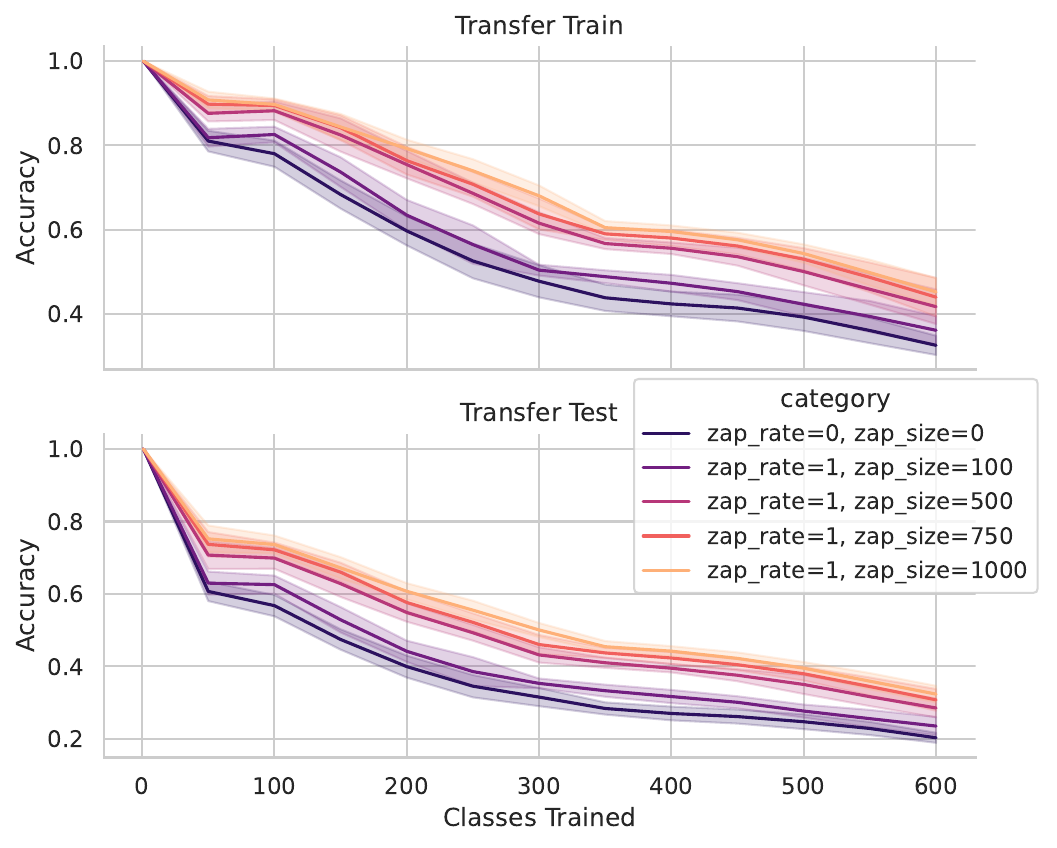}
        \label{fig:omniglot-seq-zap-rate}
    }
    \subfigure[Omniglot i.i.d.\ Transfer.]{
        \includegraphics[width=0.43\linewidth]{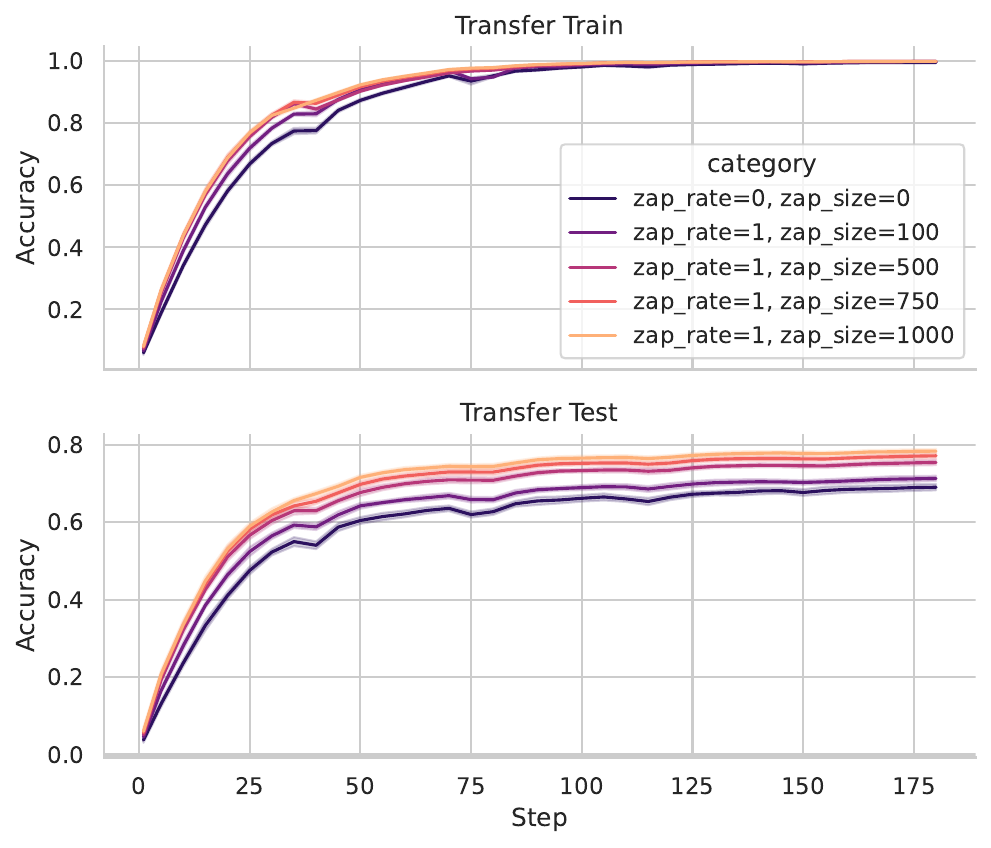}
        \label{fig:omniglot-iid-zap-rate}
    }
    \vspace{1em}
    \caption{The effect of applying different amounts of {\zapping} to standard i.i.d.\ models, pre-trained on Omniglot.}
    \vspace{1em}
    \label{fig:omniglot-zap-rate}
\end{figure}

Here we show results for all three datasets attempting different numbers of neurons for resets (differing values of $K$). We also tried resetting less often than once per epoch ($E > 1$), but resetting every epoch was typically better. In fact, in most cases below, the best scenario is to reset \textit{all} last layer weights at the beginning of \textit{every} epoch. 
Fine-tuning trajectories are shown in Figure~\ref{fig:omniglot-zap-rate} and final performance is summarized in Tables~\ref{tab:seq-zap-rate} and~\ref{tab:iid-zap-rate}.

These experiments show the potential for applying {\zapping} to standard batch gradient descent. Our variants with more {\zapping} generally see improved transfer, even though the effect is not as substantial as when it is paired with ASB. It stands to reason that there should be some point at which \textit{too} much resetting becomes detrimental, and we have not tried to reset more often than once per epoch ($E < 1$) in the i.i.d. setting, so this would be a great starting point for future work.

\begin{table}[ht]
    \begin{center}
    \begin{tabular}{c@{\hskip 0.5in}rr@{\hskip 0.5in}rr}

    \toprule
    \multicolumn{1}{c@{\hskip 0.5in}}{\bf Zap Amount}
    &\multicolumn{2}{c@{\hskip 0.5in}}{\bf Omniglot}
    &\multicolumn{2}{c}{\bf Mini-ImageNet} \\
     
    &\multicolumn{1}{c}{\bf Valid}  &\multicolumn{1}{c@{\hskip 0.5in}}{\bf Transfer}
    &\multicolumn{1}{c}{\bf Valid}  &\multicolumn{1}{c}{\bf Transfer} \\
    \midrule

    none   & 63.5 {\scriptsize $\pm 0.9$} & 20.3 {\scriptsize $\pm 1.4$} & 45.4 {\scriptsize $\pm 1.3$} &  7.4 {\scriptsize $\pm 0.8$} \\
    small  & 63.2 {\scriptsize $\pm 1.5$} & 23.5 {\scriptsize $\pm 2.5$} & 44.8 {\scriptsize $\pm 1.0$} &  7.9 {\scriptsize $\pm 1.0$} \\
    medium & 66.3 {\scriptsize $\pm 1.2$} & 21.5 {\scriptsize $\pm 6.6$} & 47.3 {\scriptsize $\pm 1.0$} &  7.8 {\scriptsize $\pm 0.7$} \\
    large  & 66.7 {\scriptsize $\pm 1.1$} & 22.0 {\scriptsize $\pm 6.5$} & 47.5 {\scriptsize $\pm 0.3$} &  7.3 {\scriptsize $\pm 0.8$} \\
    all    & 67.1 {\scriptsize $\pm 1.2$} & \textbf{32.3 {\scriptsize $\pm 2.3$}} & 47.5 {\scriptsize $\pm 0.9$} &  7.7 {\scriptsize $\pm 0.9$} \\

    \bottomrule
    \end{tabular}
    \end{center}
    \caption{Accuracy for unfrozen sequential transfer on i.i.d.\ pre-trained models with different amounts of {\zapping}. Results are aggregated in the same way as the tables in the main text.}
    \label{tab:seq-zap-rate}
\end{table}

\begin{table}[ht]
    \begin{center}
    \begin{tabular}{c@{\hskip 0.5in}rr@{\hskip 0.5in}rr}

    \toprule
    \multicolumn{1}{c@{\hskip 0.5in}}{\bf Zap Amount}
    &\multicolumn{2}{c@{\hskip 0.5in}}{\bf Omniglot}
    &\multicolumn{2}{c}{\bf Mini-ImageNet} \\
    
    &\multicolumn{1}{c}{\bf Valid}  &\multicolumn{1}{c@{\hskip 0.5in}}{\bf Transfer}
    &\multicolumn{1}{c}{\bf Valid}  &\multicolumn{1}{c}{\bf Transfer} \\
    \midrule

    none   & 63.5 {\scriptsize $\pm 0.9$} & 69.0 {\scriptsize $\pm 0.8$} & 45.4 {\scriptsize $\pm 1.3$} & 34.4 {\scriptsize $\pm 1.3$} \\
    small  & 63.2 {\scriptsize $\pm 1.5$} & 71.3 {\scriptsize $\pm 0.7$} & 44.8 {\scriptsize $\pm 1.0$} & 35.2 {\scriptsize $\pm 1.1$} \\
    medium & 66.3 {\scriptsize $\pm 1.2$} & 75.4 {\scriptsize $\pm 0.7$} & 47.3 {\scriptsize $\pm 1.0$} & \textbf{35.7 {\scriptsize $\pm 1.2$}} \\
    large  & 66.7 {\scriptsize $\pm 1.1$} & 77.1 {\scriptsize $\pm 0.7$} & 47.5 {\scriptsize $\pm 0.3$} & \textbf{36.1 {\scriptsize $\pm 0.8$}} \\
    all    & 67.1 {\scriptsize $\pm 1.2$} & \textbf{78.3 {\scriptsize $\pm 0.4$}} & 47.5 {\scriptsize $\pm 0.9$} & \textbf{36.5 {\scriptsize $\pm 1.1$}} \\

    \bottomrule
    \end{tabular}
    \end{center}
    \caption{Accuracy for standard fine-tuning on i.i.d.\ pre-trained models with different amounts of {\zapping}. Results are aggregated in the same way as the tables in the main text.}
    \label{tab:iid-zap-rate}
\end{table}

\clearpage
\section{Omni-image Dataset}
\label{si:omnimage}

\begin{figure}[!hb]
    \centering
    \includegraphics[width=0.795\linewidth]{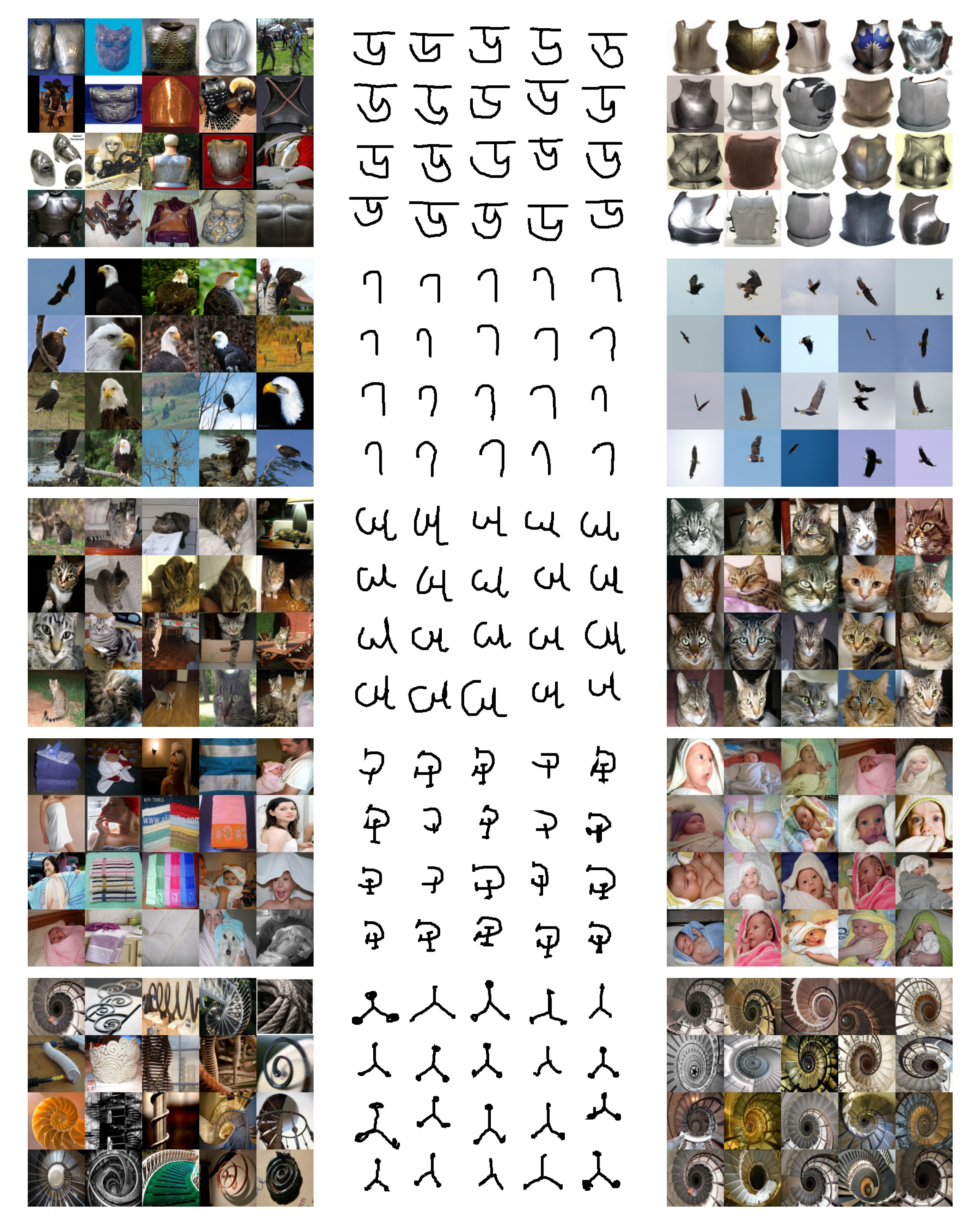}
    \vspace{1em}
    \caption{Comparing Mini-ImageNet, Omniglot and Omni-image: Samples drawn from the Mini-Image dataset (\textit{left column}) are both more complex and more visually varied than the simple and consistent one from Omniglot (\textit{center column}). On the other hand, images from Omni-image (\textit{right column}) are selected for visual similarity and more closely resemble the many classes with few-consistent examples structure of Omniglot.}
    \vspace{1em}
    \label{SI:mini_omni_image}
\end{figure}

While Mini-ImageNet contains more challenging imagery, it has far fewer classes than Omniglot, and thus cannot form a very long continual learning trajectory. To better test the effect of {\zapping} in a continual learning setting that uses natural images, we test our models on a different subset of ImageNet with a shape similar to Omniglot (1000 classes, 20 images per class), called \textit{Omni-image} \citep{omnimage23}. 
Similarly to the Omniglot case where each class contains very similar characters, the classes in Omni-image are selected to maximize within-class consistency by selecting the 20 most similar images per class via an evolutionary algorithm. See Figure~\ref{SI:mini_omni_image} for a comparison of the datasets used, highlighting the similarity between Omniglot and Omni-image, and see \citet{omnimage23}, for full details of the dataset.

\section{Unfrozen Continual Learning Results} \label{sec:seq-transfer}

\begin{figure*}[!htb]
    \centering
    \includegraphics[width=\linewidth]{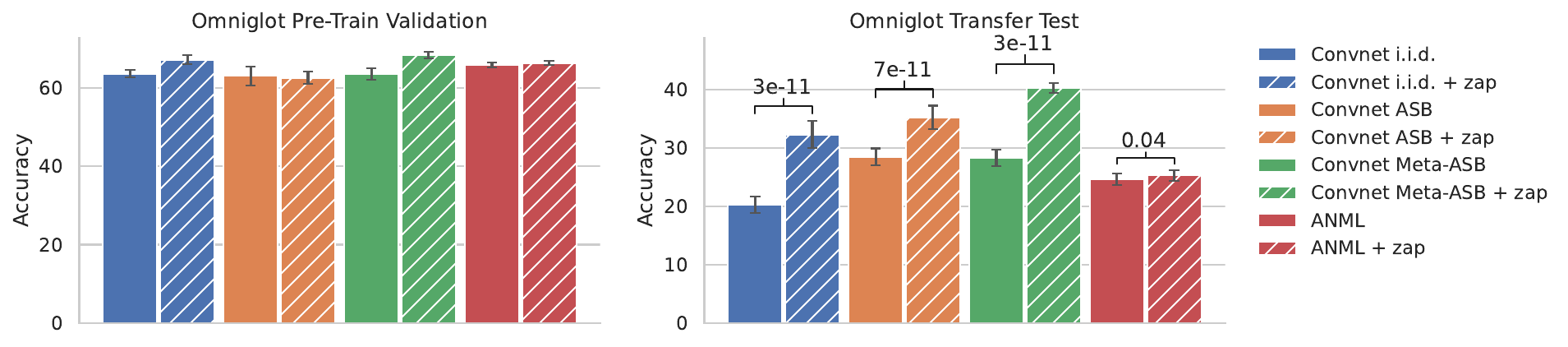} \\
    \vspace{1em}
    \includegraphics[width=\linewidth]{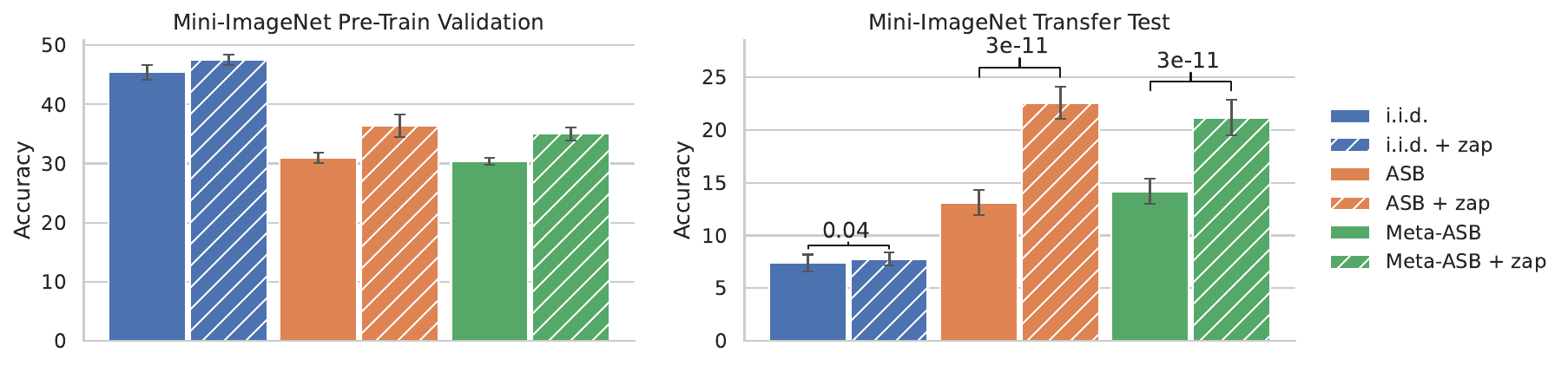} \\
    \vspace{1em}
    \caption{Average accuracy (and std dev error bars) for the unfrozen sequential transfer learning problem. 
    \textbf{Pre-Train} is the final validation accuracy of the model on the \textit{pre-training} dataset.
    \textbf{Transfer} is the accuracy on held-out instances from the \textit{transfer-to} dataset at the very end of sequential fine-tuning (i.e.\ after training on all 600 or 20 classes).
    Models trained \textbf{with {\zapping}} produce significantly ($p < 0.05$) better transfer accuracy than their counterparts without {\zapping} in all cases (p-values of a two-sided Mann-Whitney U test are shown above each pair of bars).
    Additionally, our {\sanml} architecture improves substantially over the {\anml} method when all weights are unfrozen (green vs. red in the top right plot).}
    \vspace{2em}
    \label{fig:seq-bars}
\end{figure*}

Here we show results of an \textbf{\textit{``unfrozen''} sequential transfer task}, where \textit{all} model weights are allowed to update (rather than just a linear probe). This can be compared to the ANML-Unlimited \& ANML-FT:PLN models from \citet[SI, Figure S8]{beaulieu2020learning}. This is the same as the sequential transfer described in Section~\ref{sec:evaluation}, except the entire models are fine-tuned (no weights are frozen). 

\begin{figure}[ht]
    \centering
    \includegraphics[width=0.55\linewidth]{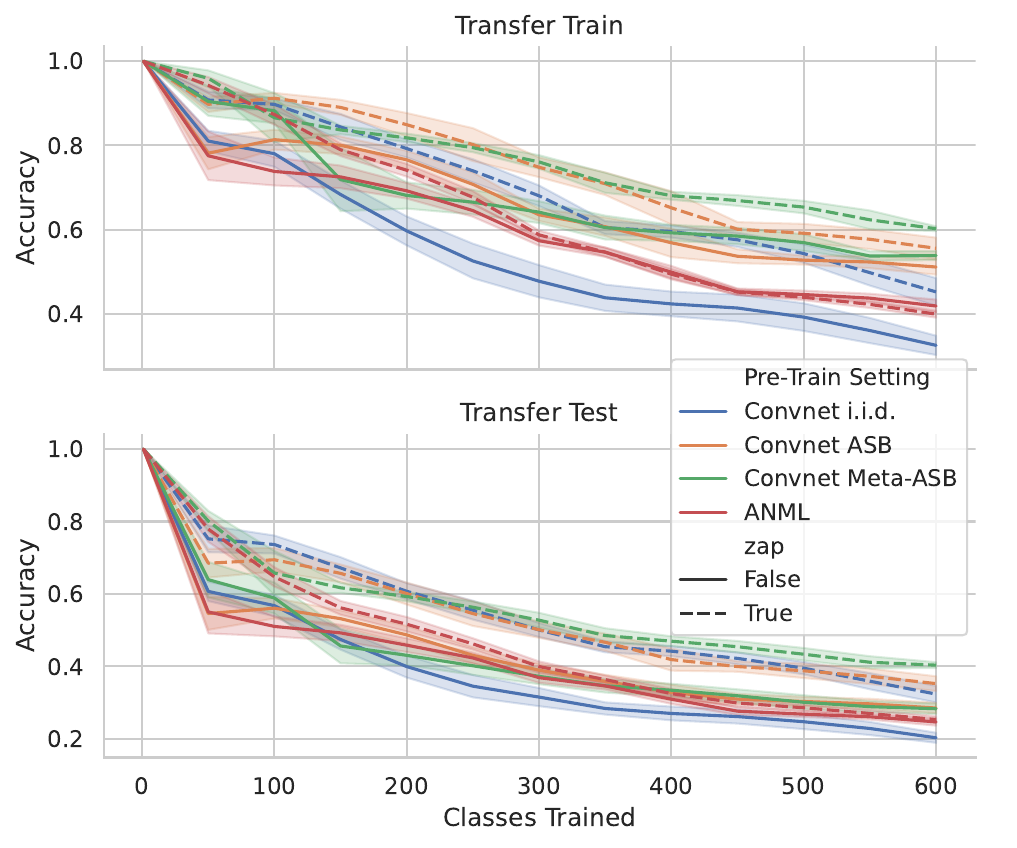}
    \vspace{1em}
    \caption{\textbf{Unfrozen Omniglot.} Accuracy on classes seen so far during unfrozen sequential transfer learning on the Omniglot dataset. \textbf{Top:} Models are trained on a few (15) examples from 600 new classes not seen during pre-training. All 15 images from a class are shown sequentially one at a time before switching to the next class. 
    \textbf{Bottom:} After each set of 50 classes (750 images), validation accuracy on the transfer set is measured using the remaining 5 examples from all classes seen up to that point.
    Models \textbf{with {\zapping}} for all three pre-training methods (dashed lines) retain significantly more validation accuracy during and after the 600 classes (9000 updates), 
    relative to models without {\zapping} (solid lines).
    }
    \label{fig:omniglot-seq}
\end{figure}

\begin{figure}[ht]
    \centering
    \includegraphics[width=0.55\linewidth]{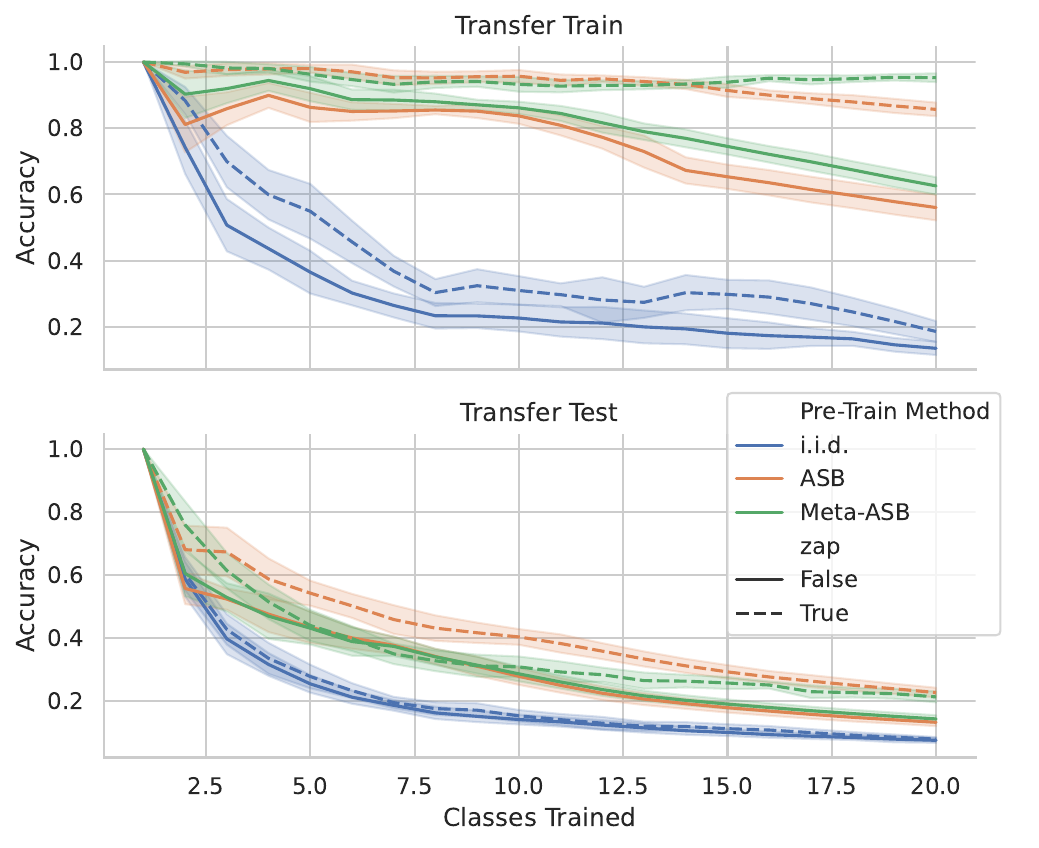}
    \vspace{1em}
    \caption{\textbf{Unfrozen Mini-ImageNet.} Accuracy on classes seen so far during unfrozen continual transfer learning on Mini-ImageNet. 
    \textbf{Top:} Models are trained on 30 examples from 20 new classes not seen during pre-training. All 30 images from a class are shown sequentially one at a time before switching to the next class. 
    \textbf{Bottom:} After each class, validation accuracy on the transfer set is measured using 100 examples per class, from all classes seen up to that point.
    Models \textbf{with {\zapping}} pre-trained with ASB (with or without meta-gradients) significantly outperform those configuration when trained without {\zapping}.  
    Models with i.i.d. pre-training show only transient improvements from {\zapping}, achieving similar train and test accuracies at the end of these training sequences, and fail to reach the final accuracies found by any of the ASB methods. 
    The training accuracy is particularly notable, where the meta+zap model is able to retain nearly 100\% of its training performance over 20 classes (600 image presentations), while the non-{\zapping} models end up around 60\%.
    }
    \label{fig:mini-imagenet-seq}
\end{figure}

Results are summarized in Figure~\ref{fig:seq-bars} and full continual learning trajectories are shown in Figures~\ref{fig:omniglot-seq} and~\ref{fig:mini-imagenet-seq}.
We see similar results to the frozen sequential transfer, in that:
\begin{itemize}
    \item models with {\zapping} significantly outperform their non-{\zapping} counterparts,
    \item ASB pre-training outperforms i.i.d.\ pre-training,
    \item and pre-train validation performance is not well-correlated with downstream sequential transfer performance.
\end{itemize}

One major difference here is that \textcolor{C3}{ANML-Unlimited} (the \textcolor{C3}{red} bars/lines) performs particularly poorly compared to our \textcolor{C2}{Convnet4} architecture, and does not benefit from {\zapping}. In this setting, Convnet4 not only saves resources but also exhibits less catastrophic forgetting. Future work to determine why {\zapping} doesn't help ANML-Unlimited may bring helpful insights to better understand the mechanics of {\zapping}.

\begin{table}[htbp]
    \begin{center}
    \begin{tabular}{cc@{\hskip 0.2in}rr@{\hskip 0.2in}rr}

    \toprule
    \multicolumn{1}{c}{\bf Pre-Train Method}
    &\multicolumn{1}{c@{\hskip 0.2in}}{\bf Zap}
    &\multicolumn{2}{c@{\hskip 0.2in}}{\bf Omniglot}
    &\multicolumn{2}{c}{\bf Mini-ImageNet} \\
    & 
    &\multicolumn{1}{c}{\bf Pre-Train}  &\multicolumn{1}{c@{\hskip 0.2in}}{\bf Transfer}
    &\multicolumn{1}{c}{\bf Pre-Train}  &\multicolumn{1}{c}{\bf Transfer} \\
    \midrule

    i.i.d.\    &      & 63.5 {\scriptsize $\pm 0.9$} & 20.3 {\scriptsize $\pm 1.4$} & 45.4 {\scriptsize $\pm 1.3$} &  7.4 {\scriptsize $\pm 0.8$} \\
    i.i.d.\    & \greencheck    & 67.1 {\scriptsize $\pm 1.2$} & 32.3 {\scriptsize $\pm 2.3$} & 47.5 {\scriptsize $\pm 0.9$} &  7.8 {\scriptsize $\pm 0.6$} \\
    ASB        &      & 63.0 {\scriptsize $\pm 2.4$} & 28.5 {\scriptsize $\pm 1.4$} & 31.0 {\scriptsize $\pm 0.9$} & 13.1 {\scriptsize $\pm 1.2$} \\
    ASB        & \greencheck    & 62.5 {\scriptsize $\pm 1.6$} & 35.2 {\scriptsize $\pm 2.0$} & 36.3 {\scriptsize $\pm 1.9$} & \textbf{22.6} {\scriptsize $\pm 1.5$} \\
    Meta-ASB   &      & 63.5 {\scriptsize $\pm 1.5$} & 28.3 {\scriptsize $\pm 1.4$} & 30.4 {\scriptsize $\pm 0.6$} & 14.2 {\scriptsize $\pm 1.2$} \\
    Meta-ASB   & \greencheck    & 68.2 {\scriptsize $\pm 0.9$} & \textbf{40.3} {\scriptsize $\pm 0.8$} & 35.0 {\scriptsize $\pm 1.1$} & \textbf{21.2} {\scriptsize $\pm 1.7$} \\
    ANML-Unlimited   & \greencheck    & 66.2 {\scriptsize $\pm 0.5$} & 25.3 {\scriptsize $\pm 0.9$} & - & - \\

    \bottomrule
    \end{tabular}
    \end{center}
    \caption{Average accuracy ($\pm$ std dev) for the unfrozen sequential transfer learning problem. 
    \textbf{Pre-Train} is the final validation accuracy of the model on the \textit{pre-training} dataset.
    \textbf{Transfer} is the accuracy on held-out instances from the \textit{transfer-to} dataset at the very end of sequential fine-tuning (i.e.\ after training on all 600/20/300 classes).
    Additionally, our {\sanml} architecture improves substantially over the {\anml} method when all weights are unfrozen.
    }
    \label{tab:unfrozen}
\end{table}

\end{document}

%% file: header.tex
\usepackage[utf8]{inputenc} 
\usepackage{hyperref}       
\usepackage{url}            

\usepackage{amsfonts}       
\usepackage{amssymb}
\usepackage{amsmath}

\usepackage{graphicx}
\usepackage{booktabs}
\usepackage{tablefootnote}
\usepackage{csquotes}
\usepackage{paralist} 
\usepackage{enumitem}

\usepackage{algorithm}
\usepackage{algorithmic}

\usepackage{dblfloatfix} 

\usepackage{subfigure}

\usepackage{xcolor}
\definecolor{snippetgray}{rgb}{.95,.95,.95}
\definecolor{eggyellow}{HTML}{e8b323}
\definecolor{viridis0}{HTML}{fde725}
\definecolor{viridis1}{HTML}{5ec962}
\definecolor{viridis2}{HTML}{21918c}
\definecolor{viridis3}{HTML}{3b528b}
\definecolor{viridis4}{HTML}{440154}
\definecolor{C0}{HTML}{1f77b4} 
\definecolor{C1}{HTML}{ff7f0e} 
\definecolor{C2}{HTML}{2ca02c} 
\definecolor{C3}{HTML}{d62728} 
\definecolor{C4}{HTML}{9467bd} 
\definecolor{C5}{HTML}{8c564b} 
\definecolor{C6}{HTML}{e377c2} 
\definecolor{C7}{HTML}{7f7f7f} 
\definecolor{C8}{HTML}{bcbd22} 
\definecolor{C9}{HTML}{17becf} 
\definecolor{egg}{HTML}{edd70c} 
\usepackage{pifont} 
\newcommand{\greencheck}{\textcolor{green}{\ding{52}}}
\newcommand{\redcross}{\textcolor{red}{\ding{53}}}

\usepackage{textcomp}
\newcommand{\mytilde}{\raisebox{0.5ex}{\texttildelow}}

\usepackage{listings}
\usepackage{realboxes} 

\newcommand{\sanml}{\textcolor{black}{Convnet}}
\newcommand{\anml}{\textcolor{black}{ANML}}
\newcommand{\oml}{\textcolor{black}{OML}}
\newcommand{\nm}{\textbf{\textcolor{red}{NM}}}
\newcommand{\rln}{\textbf{\textcolor{blue}{rln}}}
\newcommand{\pln}{\textbf{\textcolor{eggyellow}{pln}}}
\newcommand{\meta}{\textcolor{teal}{\textbf{meta}}}
\newcommand{\zap}{\textcolor{teal}{\textbf{zap}}}
\newcommand{\zapping}{zapping}
\newcommand{\Zapping}{Zapping}

\usepackage{ifthen}
\newcommand{\truefalse}[1]{\ifthenelse{\equal{#1}{true}}{\greencheck}{\redcross}}
\newcommand{\isdashed}[1]{\ifthenelse{\equal{#1}{dashed}}{-\,-}{---}}

\usepackage{marvosym}  